\documentclass[lettersize,journal]{IEEEtran}
\usepackage{amsmath,amsfonts}
\usepackage{algorithmic}
\usepackage{algorithm}
\usepackage{array}
\usepackage[caption=false,font=normalsize,labelfont=sf,textfont=sf]{subfig}
\usepackage{textcomp}
\usepackage{stfloats}
\usepackage{url}
\usepackage{verbatim}
\usepackage{graphicx}
\usepackage{cite}
\usepackage{booktabs}
\usepackage{multirow}
\usepackage[dvipsnames]{xcolor}
\usepackage{textcomp}
\definecolor{darkgreen}{RGB}{0,100,0}

\begin{document}

\title{Hierarchical Context Transformer for Multi-level Semantic Scene Understanding}
\author{Luoying Hao, Yan Hu, Yang Yue,  Li Wu, Huazhu Fu \IEEEmembership{Senior Member, IEEE}, Jinming Duan, and Jiang Liu \IEEEmembership{Senior Member, IEEE}
\thanks{Luoying Hao and Yan Hu contributed equally to this work. Corresponding author: Yan Hu, Jinming Duan and Jiang Liu.}
\thanks{Luoying Hao, Yan Hu and Jiang Liu are with the Research Institute of Trustworthy Autonomous Systems and Dept. of Computer Science and Engineering, Southern University of Science and Technology, China. (e-mail: \{huy3, liuj\}@sustech.edu.cn). Luoying Hao, Yang Yue and Jinming Duan are with the School of Computer Science, University of Birmingham, UK. Jinming Duan is also with the Division of Informatics, Imaging and Data Sciences, School of Health Sciences, University of Manchester, UK (e-mail:jinming.duan@manchester.ac.uk). Li Wu is with the MGI Tech Co., Ltd., China. Huazhu Fu is with the Institute of High Performance Computing (IHPC), Agency for Science, Technology and Research (A*STAR), Singapore. } 
}

\markboth{}%
{HAO \MakeLowercase{\textit{et al.}}: Hierarchical Context Transformer for Multi-level Semantic Scene Understanding}

\IEEEpubid{\begin{minipage}{\textwidth}\ \centering
		Copyright \copyright 20xx IEEE. Personal use of this material is permitted. \\
		However, permission to use this material for any other purposes must be obtained 
		from the IEEE by sending an email to pubs-permissions@ieee.org.
\end{minipage}}

\maketitle

\begin{abstract}
A comprehensive and explicit understanding of surgical scenes plays a vital role in developing context-aware computer-assisted systems in the operating theatre. However, few works provide systematical analysis 
to enable hierarchical surgical scene understanding. In this work, we propose to represent the tasks set [phase recognition $\rightarrow$ step recognition $\rightarrow$ action and instrument detection] as multi-level semantic scene understanding (MSSU). For this target, we propose a novel hierarchical context transformer (HCT) network and thoroughly explore the relations across the different level tasks. 
Specifically, a hierarchical relation aggregation module (HRAM) is designed to concurrently relate entries inside 
multi-level interaction information and then augment task-specific features.  To further boost the representation learning of the different tasks, inter-task contrastive learning (ICL) is presented to guide the model to learn task-wise features via absorbing complementary information from other tasks. Furthermore, considering the computational costs of the transformer, we propose HCT+ to integrate the spatial and temporal adapter to access competitive performance on substantially fewer tunable parameters. 
Extensive experiments on our cataract dataset and a publicly available endoscopic PSI-AVA dataset demonstrate the outstanding performance of our method, consistently exceeding the state-of-the-art methods by a large margin.  The code is available at https://github.com/Aurora-hao/HCT.


\end{abstract}


\section{Introduction}
\label{sec:introduction}
\begin{figure}[thp]
\begin{center}
\includegraphics[width=\linewidth]{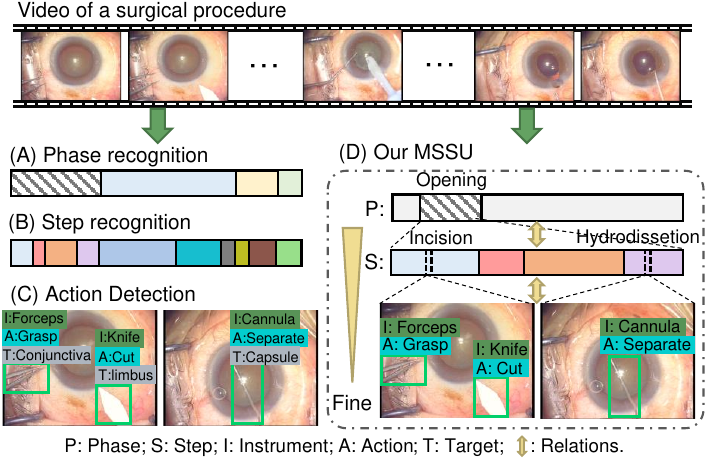}
\caption{Comparison of our proposed method with existing methods. MSSU: multi-level semantic scene understanding in surgery.}
\label{intro}
\end{center} 
\end{figure}
\IEEEPARstart{C}{Computer} assisted surgical  (CAS) systems now have a vital role in enhancing the quality of interventional healthcare \cite{vercauteren2019cai4cai}, surgeon training as well as assist procedure planning and retrospective analysis \cite{lalys2014surgical,hao2023act}. To achieve it, CAS  systems need to make use of a complete and explicit understanding of surgical scenes \cite{huaulme2021micro,lalys2014surgical, seenivasan2023surgicalgpt, wang2023truncate}.   Specifically, surgical phases and steps can monitor surgical processes and provide early alerts of potential forthcoming anomalies \cite{huaulme2020offline, zhang2023surgical}. Detailed instruments and actions appearing in the surgical scenes are beneficial to reducing preventable operation errors \cite{mascagni2022artificial, sun2023masked}, clinical decision support \cite{padoy2019machine}  and skill assessment \cite{liu2021towards, zhou2023hierarchical, li2024continual, yue2023perceptual}. 

A surgical procedure can be decomposed at different levels of granularity, such as the whole procedure, phases, steps, and actions \cite{lalys2014surgical}. 

Most of the currently available methods for surgical understanding are only focused on one specific level of granularity, like phase or step recognition at coarse granularity \cite{jin2021temporal,czempiel2020tecno,yi2022not,pan2023temporal,twinanda2016endonet,czempiel2021opera,ding2022exploring,yue2023cascade} or action recognition at fine granularity \cite{nwoye2022rendezvous,lin2022instrument, xi2022forest, wang2023vision, liu2024xfmp}, as shown in Fig. \ref{intro} (A), (B), (C) respectively. Nevertheless, a comprehensive surgical scene understanding plays a key role in developing CAS  systems, which needs to be started from a global overview of a description of high-level tasks to fine details \cite{lalys2014surgical}. Specifically, phases and steps as coarse parts provide general information about the ongoing surgical procedure, while fine-grained action and instrument detection provide a detailed perception and understanding of surgical scenes, as shown in Fig. \ref{intro} (D).
Semantic information from different levels can complement each other, forming a complete surgical portrait from various views.   However,  few works provide systematical analysis across all the levels of surgical interpretations, and those that do often fail to explore the relationships between different levels \cite{valderrama2022towards}. More critically, different levels of representation facilitate each other by  utilizing the containment information from these hierarchies.  
For  \IEEEpubidadjcol example, in cataract surgery, as shown in Fig. \ref{intro} (D), the instrument ``incision knife" is often used to act ``cut corneal-limbus", and it generally appears in phase ``opening" and step ``incision". Similarly, step ``hydrodissetion" often appears during phase ``opening" and also indicates the occurrence of both instrument ``cannula" and action ``separate capsule". Stressing on their intrinsic complementarity, the relationship between multi-level activities is essential for understanding the scene dynamics \cite{valderrama2022towards, zhang2022unsupervised}.  Therefore,  we propose a hierarchical context network for multi-level semantic scene understanding (MSSU) (phase recognition $\rightarrow$ step recognition $\rightarrow$ action and instrument detection) and explore the relationship among them.

To fully utilize the cross-task knowledge, we design an inter-task contrastive learning algorithm to further boost our model performance. Specifically, contrastive learning can automatically
learn the semantic alignment relation across different tasks according to the actual data distribution \cite{lou2023min}. In this work, we treat the inter-task representations of the same scenes as positive, while the representations of other random scenes as negative. Based on this, the contrastive learning approach is capable of maximizing the mutual information between different views of the same scene, resulting in more compact and better feature representations.

Furthermore, vision transformers have shown outstanding capabilities in visual feature representation. 
Nonetheless, the growing number of parameters and model size pose limitations on training from scratch and the extensive adoption of transformers, particularly when training video models in surgical scenes.
Parameter-efficient transfer learning is an excellent way to overcome this drawback  \cite{pan2022st,yang2023aim}.
Besides, inferring spatial and temporal structured information is a crucial aspect in the context of surgical video understanding. 
Therefore,  with the above observation, we propose a simple Spatial-Temporal Adapter  (ST-Ada) capable of learning the representations in space and time to achieve superior video understanding in surgical scenes at a small parameter cost.

In this paper,  we present a hierarchical context transformer (HCT) network to achieve MSSU in surgical videos. To realise this, we first design a hierarchical relation aggregation module (HRAM) to generate hierarchical representations for different tasks and further capture the relations between them. Then, after fusing the cross-task information guided by correlations, we obtain task-specific representations. Next, to enhance the consistency of inherently existing crossing the different tasks, we propose inter-task contrastive learning (ICL) to make the task-specific representation more discriminative. Considering the computation cost and model storage when transferring knowledge from a pre-trained model in transformer architectures, we present HCT+ network to add ST-Ada to the HCT to effectively learn spatial, and more importantly temporal representations with much fewer parameters for our multi-task learning at a significantly smaller computational cost.

 In summary, the key contributions of this work are four-fold:
 
\begin{itemize} 
\item[$\bullet$]  We propose a novel hierarchical context transformer (HCT) network to realize the multi-level semantic scene understanding MSSU and explore inter-relations between different tasks fully. MSSU represents the recognition and detection of phase $\rightarrow$ step $\rightarrow$ action and instruments in surgery.
\item[$\bullet$] We devise a  hierarchical relation aggregation module (HRAM) to encode the phase-step-instrument-action hierarchical relationship. Furthermore, for any given task, HRAM possesses the capability to aggregate information contributed by the other three tasks and exploit it to enhance task-wise representation learning. 
 
\item[$\bullet$] We develop an inter-task contrastive learning (ICL) to further strengthen the context learning capability of the model by the inherent consistency across the tasks. Considering the high computational costs of the transformer model, we integrate a  Spatial-Temporal  Adapter (ST-Ada)  into our HCT denoted as HCT+, which attains strong transfer learning abilities in spatial and temporal representations at a small parameter cost.
\item[$\bullet$] We conduct experiments with the proposed network on two datasets, our cataract video dataset and the public endoscopic PSI-AVA  dataset, to demonstrate its state-of-the-art (SOTA) performance and the contribution of each of our proposed components. The code is available at https://github.com/Aurora-hao/HCT.
\end{itemize}

\begin{figure*}[th]
\begin{center}
\includegraphics[width=\linewidth]{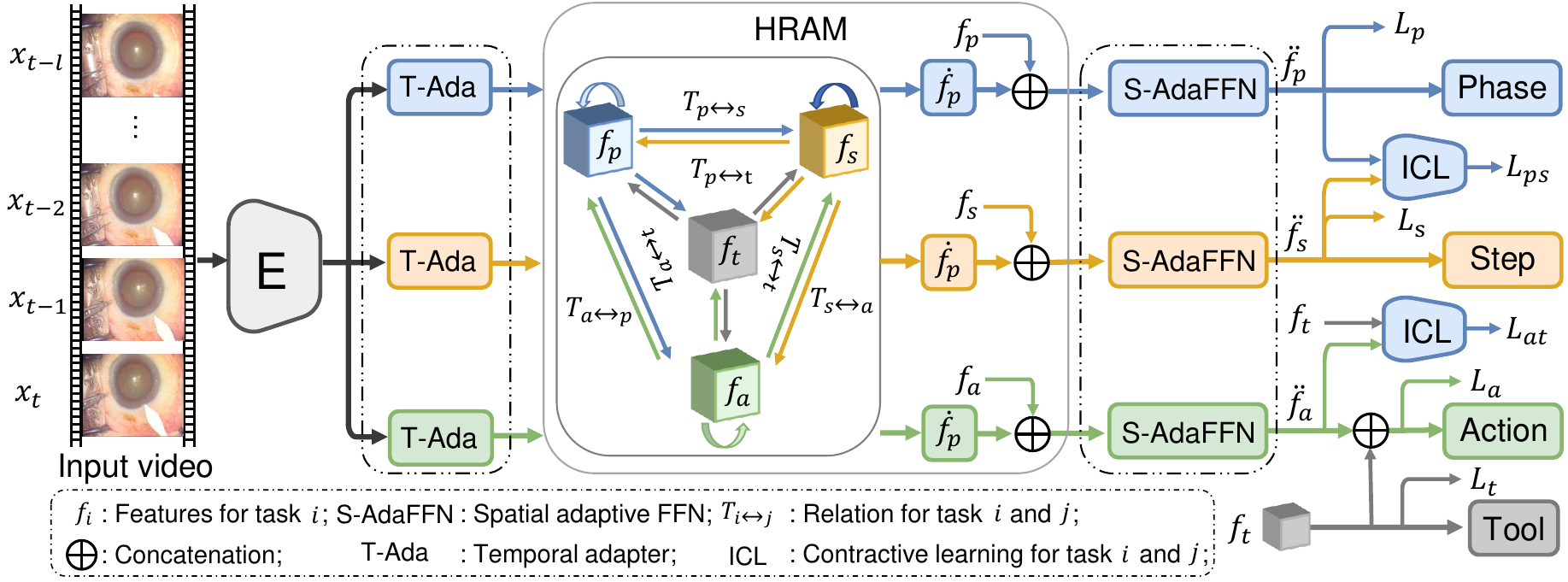}
\caption{The pipeline of our hierarchical context transformer (HCT) framework. Given an input video with $l+1$ frames, HCT uses a transformer model to extract shared features, which are then fed into the hierarchical relation aggregation module (HRAM) to capture the relations between the four task-wise features. After that, inter-task contractive learning (ICL) is utilized to further optimize the HCT. For the plus version of the model HCT+, we add a temporal adapter before HRAM and put the spatial adapter in the feed-forward module.
}
\label{pipeline1}
\end{center} 
\end{figure*}
\section{Related work}

\subsection{Surgical Scene Understanding} 
Recent works in surgical scene understanding can be broadly classified into two categories. One category aims to automatically identify surgical phases or steps within untrimmed surgical videos, which has received a lot of attention and is a very active area of research. TMRNet \cite{jin2021temporal} is proposed to establish a long-range memory bank to provide long-term temporal features and improve phase recognition accuracy. 
Yi et al.  \cite{yi2022not} proposed a non-end-to-end training strategy based on MS-TCN. After the transformer gained significant popularity for processing sequential data, Pan et al.  \cite{pan2023temporal} exploited the Swin transformer and LSTM networks 
to recognise phases in surgical videos. Czempiel et al. devised OperA \cite{czempiel2021opera}, a transformer-based model, to extract high-quality frames with an attention regularization loss.  Ding et al. \cite{ding2022exploring} presented SAHC for surgical phase recognition by emphasizing learning segment-level semantics to address erroneous predictions caused by ambiguous frames. Yue et al. \cite{yue2023cascade} fused the frame-level and phase-level temporal context for surgical workflow analysis. 

Another category focuses on finer-grained action recognition. Bawa et al. \cite{bawa2021saras} presented an endoscopic minimally invasive surgery dataset designed to tackle the problem of surgeon action detection, and provided baseline results with open-source methods like \cite{ren2015faster}. Hao et al. \cite{hao2023act}  presented ACTNet to get action locations, classes and confidence guided by anchors. Unlike existing methods for a single task, we design a new transformer module to learn the relations from multi-level tasks. Besides, we also model the constraints in the hierarchical information to regularize the framework with the ICL comprehensively.
\subsection{Multi-task learning in surgical scene}
For improving performance,  Ramesh et al. \cite{ramesh2021multi} proposed a multi-task MS-TCN to jointly predict the phases and steps, failing to get a complete surgical scene understanding.
Nwoye et al. proposed the action triplets (instrument, verb, target, as shown in Fig. \ref{intro} (C)) to represent the actions in a laparoscopic dataset \cite{nwoye2020recognition}, and a trainable 3D interaction space to capture the associations between the triplet components. They further developed a new model, the Rendezvous \cite{nwoye2022rendezvous}, to detect triplet components with class activation-guided and mixed attentions. After adding the location information of the instrument and target, Lin et al. \cite{lin2022instrument, lin2024instrument} presented a novel QDNet for the instrument-tissue interaction quintuple detection task in cataract videos. Valderrama et al. \cite{valderrama2022towards} introduced the PSI-AVA dataset and proposed a baseline framework for different level task achievement. However, they failed to harness and explore the interdependencies and correlations among these tasks. Although most of these works employ multi-task learning for action recognition, they are confined to the singular horizontal dimension of actions for more detailed analyses, while neglecting the complementarity of vertical information such as phases and steps.  Our work, in contrast, encompasses an integrated analysis across both horizontal and vertical dimensions in surgical scenarios.

\section{Methodology}
\subsection{Overview} 
The overall framework of HCT illustrated in Fig. \ref{pipeline1}, for achieving multi-level semantic activity recognition in surgical scenes, from phase and step recognition to finer-grained action and instrument/tool detection denoted by $p$, $s$, $a$ and $t$ respectively, consists of only transformer blocks. Specifically, to generate the feature of frame $x_t$, we extract a video clip as the model input, which contains the current frame $x_t$ and a set of its previous frames, as $x={x_{t-l},...,x_{t-1}, x_t}$, with $l+1$ frames. 
The clip is first passed forward a transformer backbone to generate shared feature maps $f_p$, $f_s$ and $f_a$. $f_t$ is instrument-specific features. For obtaining video features effectively to model the complex temporal cues of the surgical scenes, we utilize the MViTv2  \cite{li2022mvitv2} model as the backbone. 
It allows capturing simple low-level visual information at early layers and spatially coarse, but complex and high-dimensional features at deeper layers, benefiting from its multi-scale feature hierarchies.

After that, an HRAM is proposed to aggregate different tasks to dig out the dependencies among them and further get the task-specific features. Specifically, we utilize cross-attention in transformer blocks to get the relations $T_{i \leftrightarrow j}$ ($i,j \in [p,s, a,t]$ and $i\neq j$) cross the different task $i$ and $j$. Then the relations are fused with corresponding self-attention task-wise features to get $\dot{f_p}$, $\dot{f_s}$ and $\dot{f_a}$. To better preserve the information of each task, we add skip connections as illustrated in Fig. \ref{pipeline1} to get $\ddot{f_p}$, $\ddot{f_s}$ and $\ddot{f_a}$. Then, to acquire a more compact and better feature representation for different tasks, we develop the ICL approach to maximize the mutual information between different views of the same scenes.   In this work,  ${p \leftrightarrow s}$ and ${a \leftrightarrow t}$ are selected as two contrastive pairs because phase and step have strong consistencies for the same positive sample, as do action and instrument, which is also validated by our experiments. The HCT is also optimised by the supervision losses for each task  $\mathcal L_{p}$,  $\mathcal L_{s}$, $\mathcal L_{a}$ and  $\mathcal L_{t}$.

Furthermore, the extensive parameter size of the transformer model imposes significant limitations on its utility, particularly in the medical scenes. With this in mind, this paper introduces the temporal adapter (T-Ada in Fig. \ref{pipeline1}) placed before the HRAM and spatial adapter placed in the feed-forward network  (S-AdaFFN in Fig. \ref{pipeline1}) to construct  HCT+. It attains strong transfer learning abilities by only fine-tuning a small number of extra parameters while getting competitive performance. 
\begin{figure*}[th]
\begin{center}
\includegraphics[width=0.93\linewidth]{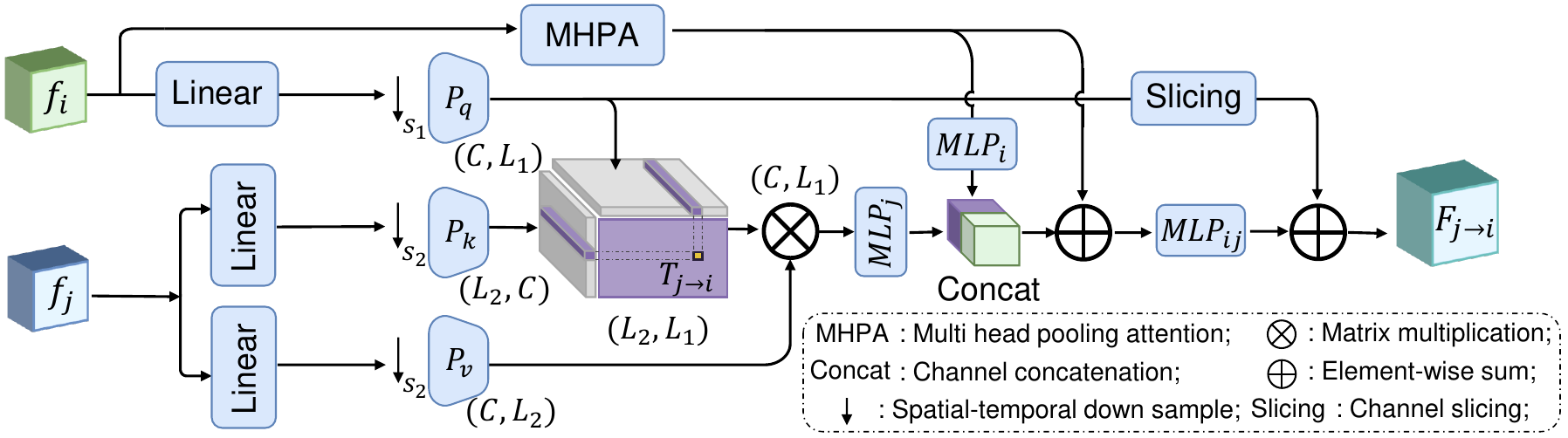}
\caption{Detailed structure of the hierarchical relation aggregation module (HRAM).}
\label{pipeline2}
\end{center}
\end{figure*}
\subsection{Hierarchical Relation Aggregation Module (HRAM)}
Transformer has strength in the computation of long-range dependencies and the general representation ability for various tasks \cite{pan2023temporal, ding2022exploring}. This advantage is essential since the single branches start from the shared features. In this work, we achieve the shared encoder built based on a stack of transformer blocks from MViTv2. After getting shared features, in order to exploit the hierarchical relationship in the surgery, HRAM is developed to generate task-specific features with complementary information inside.

As illustrated in Fig. \ref{pipeline2}, suppose there are  $n$ tasks. The input features of the HRAM are denoted as $f_i$  and $f_j$, where $i,j \in \{1,..., n\}$ and $i\neq j$. Without loss of generality, considering here task $i$ as our primary task, and task $j$ as the secondary, we seek to estimate the features $f_{j \rightarrow i}$ from task $j$ that can contribute to task $i$.  We employ two parts of attention: (i) correlation attention, which aims to exploit cross-task spatial-temporal correlations to guide the extraction of contributing features of the secondary task to the primary task; (ii) a self-attention on the primary task, to retain and further enhance information from the primary task.

Concretely,  in correlation attention part,  for $f_i, f_j\in \mathbb{R}^{L \times C}$ ($L = l \times h\times m$, $l$ denotes length of input video clip, $h$ and $m$ are the height and width of feature map respectively), it applies pooling operations to perform spatial-temporal downsampling with $s_1$ and $s_2$  scale factor respectively,  to get query $f_q$, key $f_k$ and value $f_v$ tensors \cite{li2022mvitv2}.  
\begin{equation}
\begin{split}
\nonumber
&f_q=P_q(f_iW_q;s_1) \quad f_k=P_k(f_jW_k;s_2) \quad f_v=P_v(f_jW_v;s_2),
\label{eq1}
\end{split}
\end{equation}
where $f_q \in \mathbb{R}^{L_1  \times C}$  and $f_k, f_v \in \mathbb{R}^{L_2\times C}$.  $W_q, W_k, W_v$ represent the projection matrices. Spatial-temporal correlation attention matrix $C_{j \rightarrow i}\in \mathbb{R}^{L_1 \times C}$ is then obtained :
\begin{equation}
\begin{split}
\nonumber
&C_{j \rightarrow i} = softmax(\frac{f_qf_k^{T}}{\sqrt{d}})f_v,
\label{eq1}
\end{split}
\end{equation}
where $\sqrt{d}$ is used to normalize the inner product matrix row-wise. Intuitively, $C_{j \rightarrow i}$ has high values where features from task $i$ and $j$ are highly correlated and low values otherwise.  For the self-attention part on primary task $i$, multi-head pooling attention (MHPA) \cite{fan2021multiscale} is employed to enable progressive change at the spatial-temporal resolution in the transformer backbone. 

Subsequently,  $n$  mapping functions are utilized to adjust the output dimension and fuse different parts of attention,  the integrated feature $T_{j \rightarrow i}$ for task $i$,  enhanced by the correlation between task $i$ and $j$, is defined as: 
\begin{equation}
\begin{split}
\nonumber
&T_{j \rightarrow i} =  concat \left( MLP_j(C_{j \rightarrow i}), MLP_i({\rm MHPA}(f_i))\right),
\label{eq1}
\end{split}
\end{equation}
where  $MLP_j$  is designed to transfer the size of  $C_{j \rightarrow i}$  to $ L_1 \times \frac{C}{n-1}$, while $MLP_i$  is designed to maintain the dimension of feature $f_i$.  $concat$  indicates the channel-wise concatenation. 

After that,  the concatenated feature map is then processed with the element-wise addition with the self-attention information of $f_i$,  followed by another mapping function $MLP_{ij}$, leading to the final refined features. Consequently, for aggregating all correlated secondary task information contributing towards  primary task $i$, we can get final refined task-specific features $F_{all \rightarrow i}$:
\begin{equation}
\begin{split}
\nonumber
&T_{all \rightarrow i} = MLP_{ij}\left(concat_{j=1}^n \left( T_{j \rightarrow i}\right) + {\rm MHPA}(f_i)\right),\\
&F_{all \rightarrow i} = T_{all \rightarrow i} + slicing(f_q),
\label{eq1}
\end{split}
\end{equation}
where  $slicing(\cdot)$  indicates channel slicing for achieving skip connection. The slicing operation shown in Fig. 3 is optional, and is particularly useful in instrument-specific feature computation. It ensures dimensional consistency and filters out irrelevant information, which is essential for accurate processing in the task. $MLP_{ij}$  is employed to keep the input $f_i$ and output  dimensions consistent. By design, the refined primary task features have the same dimensions as the input features.

In this paper, we intend to achieve four tasks to attain a hierarchical understanding of surgical scenes. For the tasks of phase, step recognition and action detection, we fuse relevant information from the other three tasks and get final refined task-specific features $F_{all \rightarrow i}$. To account for instrument detection, we employ the DINO \cite{zhang2023dino} to obtain bounding box estimations and box-specific features, then feed them into the HRAM module. Therefore, when we use instrument-specific features as auxiliary roles,  the architecture consisting of two linear layers and an activation layer in the middle and padding operation is proposed to keep the same dimension as the primary task features to operate following attention calculation. What is more, as shown in Fig.  \ref{pipeline1}, to achieve the action detection task, we integrate action-specific features and instrument-specific features to obtain final results.


\subsection{Inter-task Contrastive Learning (ICL)}
\begin{figure}[tb]
\begin{center}
\includegraphics[width=\linewidth]{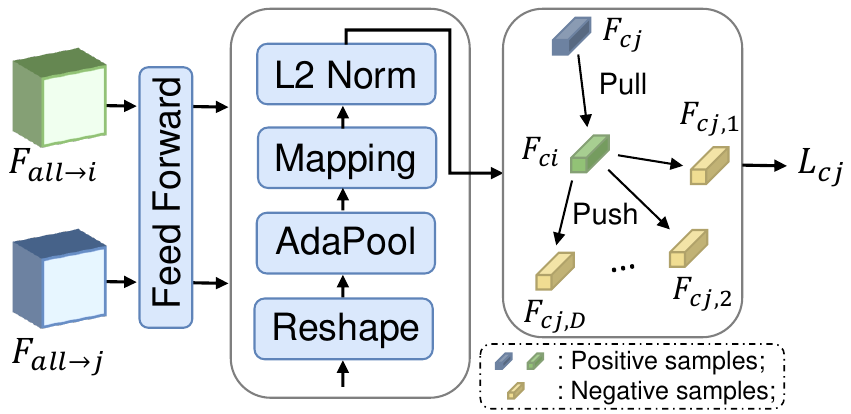}
\caption{Detailed structure of the inter-task contrastive learning (ICL).}
\label{pipeline3}
\end{center}
\end{figure}

As the different tasks are learned from the same input video clip data, consistency inherently exists crossing the different tasks in a multi-task prediction. The model performance of one task can thus benefit from utilizing the consistent information from other tasks to boost task-specific representation learning. To better illustrate such consistency, we propose ICL bring their representations closer together, which brings task pairs (positive examples) near together while pushing irrelevant ones (negative examples) far apart. As shown in Fig. \ref{pipeline3}, the distances of positive pairs are consistently smaller than those of negative ones on different target tasks. 

Specifically, with input video clips, we treat the representations from two different tasks in the same scene (same video clips) as positive pairs, while the representations in different scenes (different video clips) as negative samples. For $n$ tasks in our paper, we take two of them as examples denoted by $i$ and $j$ as task pairs for contrastive learning. We set the feature obtained from the HRAM as the input of ICL. As illustrated in Fig. \ref{pipeline3}, for the features $F_{all \rightarrow i}$ and  $F_{all \rightarrow j}$,  after the feed-forward process in the transformer block, we first average them in terms of the space-time dimension by adaptive average pooling, followed by a set of mapping function and $L_2$ normalization. Then, we get the reshaped features $F_{ci}$  and $F_{cj}$. Given a positive example of task pair, we randomly pick a set of $D$ $\{F_{cj,d}\}_{i=1}^{D}$  for $F_{ci}$  from the same batch as negative examples and vice versa. Therefore, the losses for tasks $i$ and $j$ are as follows:
\begin{equation}
\begin{split}
\nonumber
&\mathcal L_{ci} = -\sum_{F_{ci},F_{cj}}log{\frac{exp(sim(F_{ci},F_{cj})/\tau}{\sum_{d\in \Lambda}exp(sim(F_{ci},F_{cj,d})/\tau)}},\\
&\mathcal L_{cj} = -\sum_{F_{ci},F_{cj}}log{\frac{exp(sim(F_{ci},F_{cj})/\tau}{\sum_{d\in 
 \Lambda}exp(sim(F_{ci,d},F_{cj})/\tau)}},\\
&\mathcal L_{cij} = \mathcal L_{ci}+\mathcal L_{cj},
\label{eq1}
\end{split}
\end{equation}
where $\Lambda$ represents a batch of video clips, and $\tau$ is the temperature parameter that controls the strength to contrast. $sim$ is defined as $sim(x,y)=\frac{x^Ty}{\vert x \vert \vert y \vert}$ . In the training process, negative examples are from the same training batch of data. $\mathcal L_{cij}$  denotes ICL loss between task $i$ and task $j$. In experiments, phases and steps are selected to form pairs and fed into ICL due to their strong inherent consistency in the same positive samples. This holds true for pairs composed of actions and instruments as well, which is validated by subsequent ablation studies.
\begin{figure}[bhp]
\begin{center}
\includegraphics[width=0.9\linewidth]{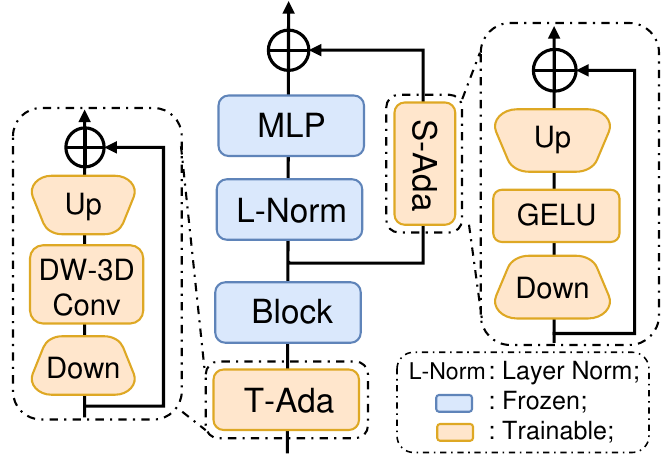}

\caption{Detailed structure of the spatial-temporal adapter in our proposed transformer block.}
\label{pipeline4}
\end{center}
\end{figure}

\subsection{HCT+ network}
Transfer learning is an efficient way to transfer knowledge from one domain to another to improve its performance. However, limited by the growing number of parameters and model size, 
fully fine-tuning the whole transformer model for every single downstream task would become prohibitively expensive and infeasible in training cost and model storage, especially for surgical video datasets \cite{pan2022st}.  In this work, we integrate a spatial-temporal adapter into our improved transformer block, which attains strong transfer learning abilities by only fine-tuning a small number of extra parameters while achieving comparable performance with the full fine-tuned model. 

Video understanding requires the model to learn both good appearance representations in each frame (spatial modeling) and also infer the temporal structured information across frames (temporal modeling). Hence, temporal and spatial adapters are employed in different places of the transformer block to model disparate information.  Inspired by the Adapter 
\cite{houlsby2019parameter} designed for parameter-efficient transfer learning in NLP,  we adopt spatial and temporal adapters \cite{pan2022st,yang2023aim}. 

As shown in  Fig. \ref{pipeline4}, before the transformer block, the input video clip features are firstly fed into the temporal adapter denoted by T-Ada, which is a bottleneck architecture that consists of two fully connected (FC) layers and a depth-wise 3D-convolution (DW-Conv3D) in the middle \cite{pan2022st}. The first FC layer projects the input to a lower dimension and the second FC layer projects it back to the original dimension. Formally, the temporal adapter $T\_Ada$ can be expressed as:
\begin{equation}
\begin{split}
\nonumber
&T\_Ada(f_i) = f_i+\mathbf{W_{up}}\left(DW\_Conv3D(\mathbf{W_{down}}f_i)\right),
\label{eq1}
\end{split}
\end{equation}
where the $\mathbf{W_{up}}\in \mathbb{R}^{L \times \hat{L}}$ and $\mathbf{W_{down}}\in \mathbb{R}^{\hat{L} \times L}$ denote the parameters of the up-projection layer and the down-projection layer, where $\hat{L}$ is the bottleneck middle dimension and satisfies $\hat{L}\textless \textless L$. 

In contrast to the temporal adapter, the spatial adapter employs an intermediate activation layer GELU instead of DW-Conv3D.  With the output features $F_{all \rightarrow i}$ of  transformer block as input, the spatial adapter operator $S\_Ada$ is formally written as:
\begin{equation}
\begin{split}
\nonumber
&S\_Ada(F_{all \rightarrow i}) = F_{all \rightarrow i}+\mathbf{W_{up}}\left(GELU(\mathbf{W_{down}}F_{all \rightarrow i})\right).
\label{eq1}
\end{split}
\end{equation}

For proper adaptation, we add the spatial adapter to the feed-forward module to transfer spatial information, denoted by S-Ada in Fig. \ref{pipeline4}. During training, all the other layers of the transformer model are frozen while only the spatial and temporal adapters are updated \cite{houlsby2019parameter}.

\subsection{Overall Training Objective }
In this work, we employ a weighted binary cross-entropy loss for action detection while a  weighted cross-entropy loss is utilized for the phase, step and instrument recognition. Furthermore, the ICL losses for task pairs are added to enhance the task-specific feature extraction. The final training objectives can be written as:
\begin{equation}
\begin{split}
\nonumber
&\mathcal L_f = \sum _{i=1}^n\lambda _i \mathcal L_i +\sum _{i,j=1,i\neq j }^n \mathcal L_{cij},
\label{eq1}
\end{split}
\end{equation}
 where $\lambda _i$, $\mathcal L_i$ denote the weights and loss functions for different tasks, and $n$ is the number of tasks.

\begin{table*}[tb]
    \centering
\caption{ Performance Comparison on the cataract and PSI-AVA dataset. The best results are highlighted in bold. ``Frames" denotes the number od frames used in models. }
\label{tabel1}
\setlength{\tabcolsep}{1.5mm}{
    \begin{tabular}{cccccccc|cccccc} 
    \toprule
        \multirow{3}*{Method}& \multirow{3}*{Frames} &\multicolumn{6}{c|}{Cataract dataset}  &\multicolumn{6}{c}{PSI-AVA dataset}\\
        \cline{3-8} \cline{9-14}
          & &   \multicolumn{2}{c}{Phase}&   \multicolumn{2}{c}{Step}&   Instrument& Action&  \multicolumn{2}{c}{Phase}&  \multicolumn{2}{c}{Step}&   Instrument& Action\\ 
          \cline{3-8} \cline{9-14}
          & &   mAP&  Acc&    mAP&  Acc&  mAP& mAP&  mAP&  Acc&    mAP&  Acc&  mAP&  mAP\\  \hline 
         TMRNet \cite{jin2021temporal}& 30& 0.9496  & 0.9358 &   0.9375& 0.9230  &  —& —  &0.5439 &0.5734 &0.4061&0.5357 & —& —\\
         NETE \cite{yi2022not}& whole video & 0.9312& 0.9356 & 0.8335 & 0.8921  &  —& — &  0.4806  & 0.6403  & 0.2956& 0.5116 & —& — \\  
 SAHC \cite{ding2022exploring}& whole video & 0.9358& 0.9289 & 0.8982& 0.9109 & —&—  & 0.5522 & 0.6436& 0.3798& 0.5272 &—\\ \hline\hline
         ACTNet \cite{hao2023act}& 16&  —&  —& —& —&  —& 0.3970  &  —&  —& —&  —& —& 0.2259\\ \hline\hline 
         SlowFast \cite{feichtenhofer2019slowfast} &  16& 0.9075&0.9153 & 0.8486&0.8983 & 0.8525& 0.4745 &  0.4680& 0.5463 & 0.3386& 0.4713 & 0.8029& 0.1857\\ 
         TAPIR \cite{valderrama2022towards}& 16& 0.9286&0.9472 &0.9191& 0.9528 & 0.8541& 0.5251 &  0.5824& 0.5996 & 0.4561 & 0.5228 & 0.8082& 0.2356\\ 
         MViTv2 \cite{li2022mvitv2}&  16& 0.9494&\textbf{0.9557} &0.9289&0.9329 & 0.8543& 0.5563 &  0.6189&  0.6361& 0.4837 & 0.5548  &  0.8206&0.2729\\ \hline
 HCT& 16& \textbf{0.9635}& 0.9468 &  \textbf{0.9581}& \textbf{0.9556} &  \textbf{0.8545} &\textbf{0.5726} & \textbf{0.6457}& \textbf{0.6603} &  \textbf{0.4977}& \textbf{0.5766} & \textbf{0.8217}&\textbf{0.2820 }\\ 
    \bottomrule   
    \end{tabular}}

\end{table*}

\section{Experiments}
\subsection{Datasets}
The effectiveness of our HCT is evaluated on two datasets,   our cataract dataset and public PSI-AVA \cite{valderrama2022towards}, which are derived from distinct surgical scenes. 
Our private cataract dataset comprises 20 videos (a total of 17511 frames) with a frame rate of 1 fps and resolution of 720 $\times$ 576. The dataset is provided with manual annotations done by surgeons indicating the surgical phase and step each video frame belongs to, and the actions and tools appearing in the scene. Specifically, the cataract videos consist of 4 phases, 10 steps,  49 actions and 13 instruments. For each video frame, the phase and step are provided with class labels, while the actions and instruments are labelled with the classes and locations (with bounding boxes). The cataract video dataset is randomly split into a training set with 15 videos (13583 frames) and a testing set with 5 videos (3928 frames). 
 
PSI-AVA dataset \cite{valderrama2022towards} includes eight radical prostatectomy surgeries, performed with the Da Vinci SI3000 Surgical System.  It includes 11 phases, 21 steps,  16 actions and 7 instruments, and  resolution is  1280 $\times$ 800. The phase and step annotations are provided by frames at 1 fps, while the actions and instruments are given by frames every 35 seconds of video.  Thus, 73,618 keyframes are annotated with one phase-step pair, while  2238 keyframes have instruments and actions annotations. We split the dataset following work \cite{valderrama2022towards}.

\subsection{Implementation Details}
The proposed architecture is built with Pytorch and is trained on NVIDIA RTX A6000. MViTv2 \cite{li2022mvitv2} pretrained on Kinetics-400 \cite{carreira2017quo} is adapted as the backbone. We use AdamW optimizer with a base learning rate of  $1e-3$, and  A cosine  scheduler with a weight decay of 0.05.
The model is totally trained for 60 epochs with a 10 epoch warm-up period. Data augmentation with 224 $\times$ 224 cropping, jittering and flipping is performed to enlarge the training dataset. The sequence length is 16 frames as \cite{li2022mvitv2}. The weights of the losses for phase, step, instrument and action tasks are set as 0.3, 0.2, 0.3 and 0.2 respectively. The batch size for both the baseline and the proposed method is consistent, set to 20. The temperature parameter in the ICL module is set to 0.07. The settings are the same for the two datasets. 
For getting the box-specific features, we pretrain DINO-5scale \cite{zhang2023dino}  
on training data for our cataract datasets. For PSI-AVA dataset, due to the limited annotation of instruments, we conduct pretraining based on the EndoVis 2017 \cite{allan20192017} and EndoVis 2018 dataset \cite{allan20202018}  following up \cite{valderrama2022towards}, and then train on PSI-AVA.  Bounding boxes are selected with a confidence threshold of 0.75, and corresponding 256-dimensional feature embeddings are extracted as box-specific features.  To assess performance in the phase and step recognitions, we use the mean Average Precision (mAP) and Accuracy (Acc), the standard metrics in videos.  For instrument and action detection, we use the mAP metric at an Intersection-over-Union threshold of 0.5 (mAP@0.5IoU). 

\begin{table}[bp]
\centering
\caption{Ablation study on key Components based on the two datasets. }
\label{tabel3}
\setlength{\tabcolsep}{2.3mm}{
    \begin{tabular}{cc|cccc}
        \toprule
        \multicolumn{2}{c|}{Modules} & \multirow{2}*{Phase}&  \multirow{2}*{Step}&  \multirow{2}*{Instrument}& \multirow{2}*{Action} \\
        \cline{1-2}
         HRAM& ICL&    \multicolumn{4}{c}{} \\ \hline
        \multicolumn{6}{c}{Cataract dataset} \\ 
        \cline{1-6}
         &  &    0.9494 &0.9289 & 0.8543& 0.5563\\
        \checkmark&  &    0.9523&  0.9489&  \textbf{0.8547}&   0.5682\\
        \checkmark &  \checkmark&    \textbf{0.9635}&  \textbf{0.9581}&  0.8545&   \textbf{0.5726}\\\hline
         \multicolumn{6}{c}{PSI-AVA dataset} \\ 
        \cline{1-6}
         &  &    0.6189&  0.4837& 0.8206& 0.2729\\
         \checkmark&  &    0.6251&  0.4914&  0.8213&   0.2794\\
         \checkmark&  \checkmark&   \textbf{0.6457}&  \textbf{0.4977}&  \textbf{0.8217}&   \textbf{0.2820}\\ \bottomrule
    \end{tabular}}
\end{table}

\subsection{Comparison with State-of-the-arts and Visualisation}
\subsubsection{Comparison with State-of-the-arts}
Table \ref{tabel1} presents a performance comparison between HCT and the SOTA methods on our cataract dataset and the public PSI-AVA dataset. The best results are highlighted in bold. 
Given that instrument detection is comparatively simpler and the current method has achieved high accuracy, our analysis is primarily focused on the other three tasks. 

To validate the performance of our model, we carry on comparisons between single-task models TMRNet \cite{jin2021temporal}, NETE \cite{yi2022not} and SAHC \cite{ding2022exploring} in phase and step recognition, as well as action detection ACTNet\cite{hao2023act}. Additionally, we perform multi-task performance comparisons based on various video recognition methods  SlowFast \cite{feichtenhofer2019slowfast},   TAPIR\cite{valderrama2022towards} and  MViTv2 \cite{li2022mvitv2} by adding the same multi-task losses and heads. 

Using the cataract dataset, TMRNet achieves the best performance among the three existing methods, indicating that longer video frames (30 frames) significantly enhance phase and step recognition. However, compared to TMRNet, our method still achieves a 1.39\%  (96.35\% vs. 94.96\%) improvement in phase recognition and a 2.06\% (95.81\% vs. 93.75\%) improvement in step tasks on mAP, even though our method only uses a sequence of 16 frames, while NETE \cite{yi2022not} and SAHC \cite{ding2022exploring} use whole video features in models. Furthermore, all three methods require at least two training stages on the training dataset, whereas our approach can achieve end-to-end recognition for both phase and step. Our model on action detection, compared with  ACTNet, an anchor-context action detection network, also achieves a significant improvement in terms of mAP.

For the multi-task effect comparison, we adopt three popular SOTA methods in action recognition, encompassing both CNN-based and transformer-based approaches. By replacing the original action recognition heads of these methods with our multi-task heads, we facilitate a comparison of multi-task effects. To ensure fairness, neither the three methods nor our approach utilizes pre-trained models. The results demonstrate that our method outperforms others in every task. Specifically, in the tasks of phase, step and action, we exceed the performance of the best comparison method MViTv2 by 1.41\% (96.35\% vs. 94.94\%), 2.92\% (95.81\% vs. 92.89\%) and 1.63\% (57.26\% vs. 55.63\%) respectively, highlighting the strong synergistic effect among these four tasks. Regarding the lower phase recognition accuracy on the cataract dataset compared to the baseline, we calculate the Balanced Accuracy (B-Acc) for both methods (MViTv2 92.84\% vs. HCT 94.54\%), as shown in the supplementary materials. Our method shows a 0.89\% lower Acc but a 1.7\% higher B-Acc than MViTv2. This indicates that while MViTv2 performs better on more frequent categories, our approach achieves a more balanced performance across all categories. Notably, it excels in more challenging categories with fewer samples, making it more suitable for practical applications.

\begin{table}[bp]
    \centering
\caption{Ablation study on HRAM. }
\label{tabel4}
\setlength{\tabcolsep}{2.2mm}{
    \begin{tabular}{cccc|cccc}
        \toprule
        \multicolumn{4}{c|}{Items} & \multirow{2}*{Phase}&  \multirow{2}*{Step}&  \multirow{2}*{Instrument}& \multirow{2}*{Action} \\
        \cline{1-4}
         P& S&  I&  A&  \multicolumn{4}{c}{} \\\hline
         &  \checkmark&  \checkmark&  \checkmark&  0.9153&  0.9235&  0.8544& 0.5497\\
         \checkmark&  &  \checkmark&  \checkmark&  0.9337&  0.8739&  0.8545& 0.5561\\
 \checkmark& \checkmark& & \checkmark& 0.9045& 0.8827& 0.8534&0.5423\\
          \checkmark&  \checkmark&  \checkmark&  &  0.9412&  0.9046&  0.8515& 0.5471\\
         \checkmark&  \checkmark&  \checkmark&  \checkmark&  
\textbf{0.9523}&  \textbf{0.9489}&  \textbf{0.8547}& 
\textbf{0.5682}\\\bottomrule
    \end{tabular}}
\end{table}

Regarding the PSI-AVA dataset, we conduct similar comparisons, as shown in Table \ref{tabel1}. 
It's important to note that the PSI-AVA dataset, compared to our cataract dataset, has considerable differences in annotations for each task. Unlike our cataract dataset, where every frame is annotated with labels or bounding boxes for phase, step, instrument, and action at 1fps, the PSI-AVA dataset annotates only phase and step labels at 1fps, with instrument and action labels annotated at 35-second intervals.  For a fair comparison,
we follow the same training approach as in existing work \cite{valderrama2022towards}. It involves first training for phase and step recognition, followed by instrument and action detection using the best-trained step recognition model. This also results in the PSI-AVA dataset not benefiting from the mutual promotion of the four tasks during training, as observed in the cataract dataset training.  For phase and step recognition tasks, the model is only optimized by the interaction between phase and step,  and similarly for instrument and action detection.  Nevertheless, our method still achieves the best performance in each task, further validating the effectiveness of our framework.

\begin{figure}[tp]
\begin{center}
\includegraphics[width=\linewidth]{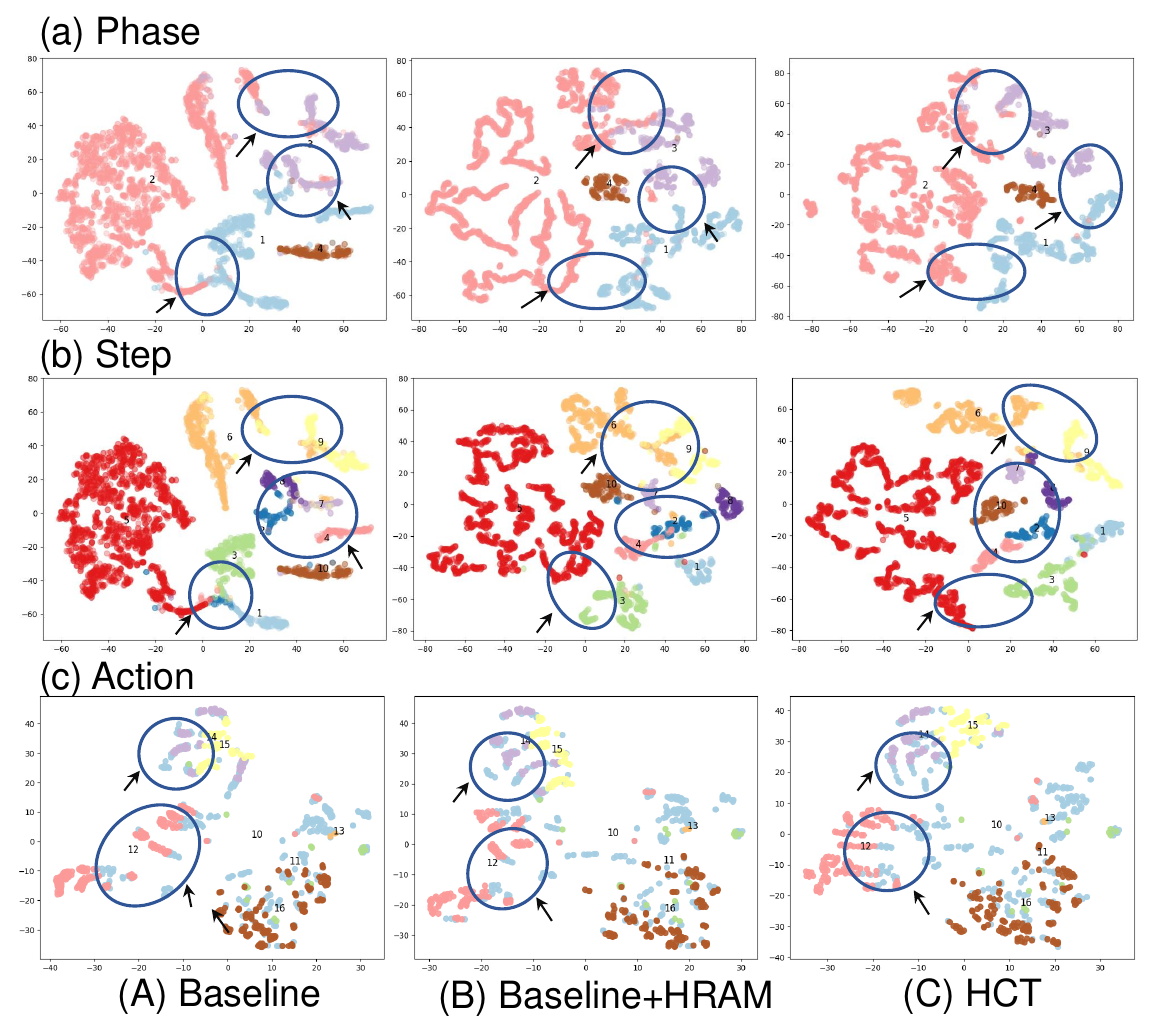}

\caption{t-SNE Visualization results of (A) baseline, (B) baseline+HRAM and (C) HCT in phase, step and action recognition task.  The blue circle indicates improvement in effectiveness. Due to space constraints, only the distribution of the 7  challenging action classes related to the instrument ``Cannula" are shown in the figures, illustrating our method's enhanced performance on these difficult action classes.}

\label{tsne}
\end{center}
\end{figure}
\begin{figure}[tp]
\begin{center}
\includegraphics[width=\linewidth]{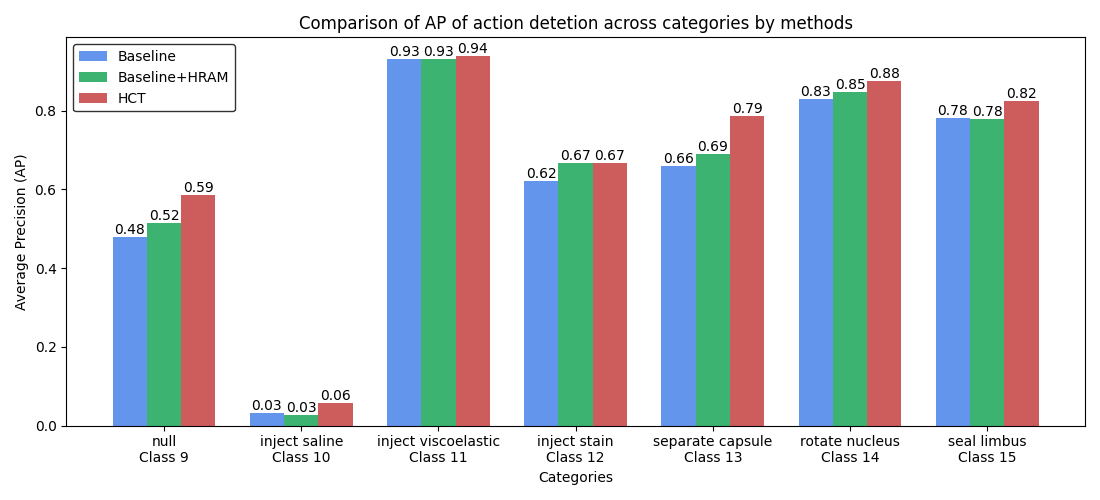}

\caption{Comparision of Average Precision (AP) of action detection across the challenging classes related to the instrument ``Cannula" in Fig. 6, to quantitatively demonstrate enhanced effectiveness.}
\label{bar}
\end{center}
\end{figure}

\begin{figure*}[t]
\begin{center}
\includegraphics[width=0.89\linewidth]{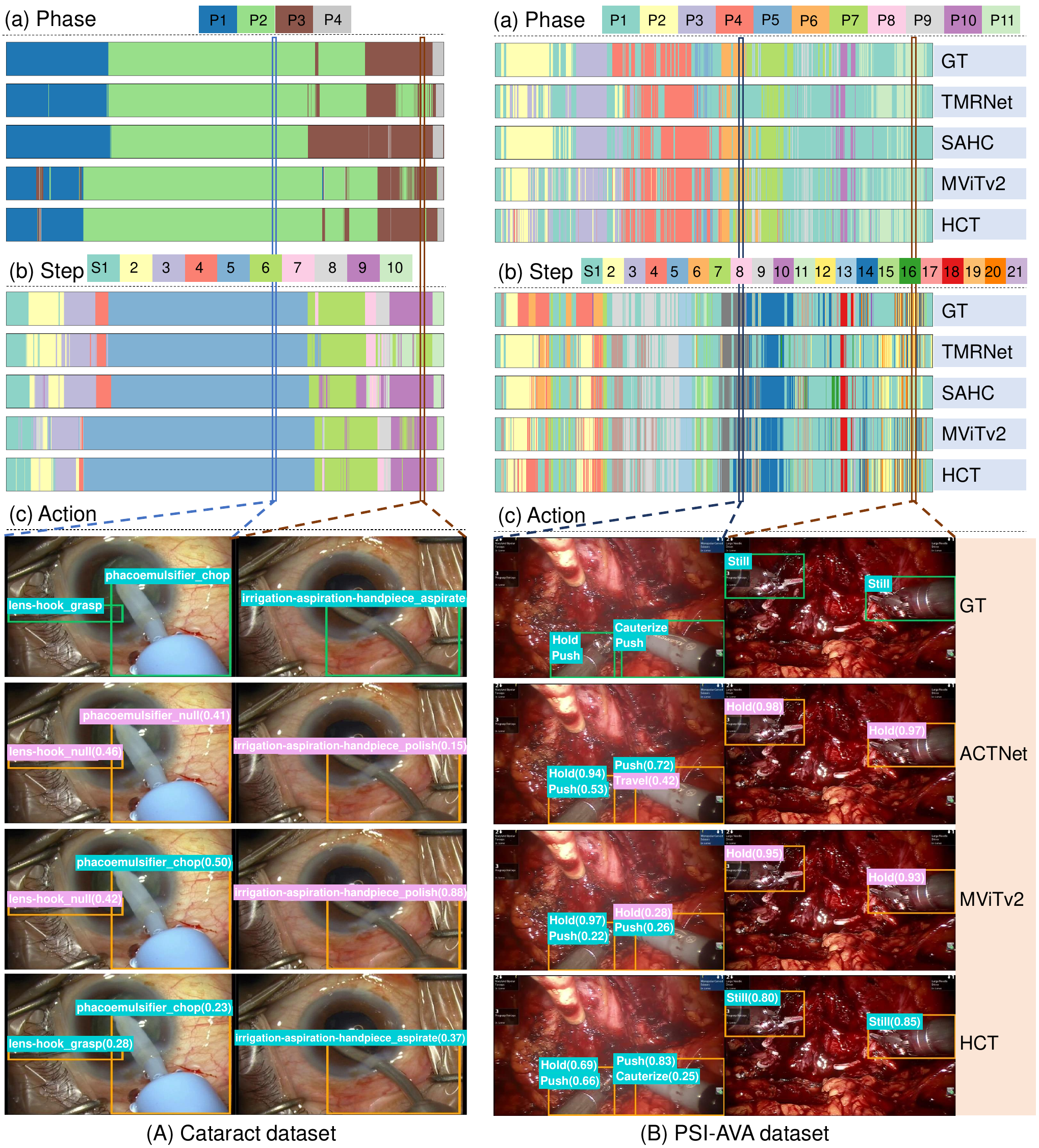}
\caption{Multi-level visualisation based on (A) Video66 from cataract dataset and (B) CASE015 from PSI-AVA dataset. (a) Qualitative comparisons for phase recognition. (b) Qualitative comparisons for step recognition. (c) Qualitative comparisons for action detection. We show the ground-truth (GT) ({\color{cyan}{sky blue}}), right predictions (\textcolor{cyan}{sky blue}) and wrong predictions ({\color{Thistle}light purple}).}
\label{result1}
\end{center}
\end{figure*}

\subsubsection{Result Visualisation and Analysis}


To illustrate the effectiveness of HCT more clearly, we list visualizations of prediction results for the two datasets in Fig. \ref{result1}, which presents the multi-level semantic understanding of surgical videos spanning coarse to fine-grained tasks.
Limited by the page, we only list the partial results with high accuracy, TMRNet \cite{jin2021temporal},  SAHC \cite{ding2022exploring} and  MViTv2 \cite{li2022mvitv2} for the phase and step recognition, while ACTNet \cite{hao2023act} and MViTv2 \cite{li2022mvitv2} for action detection. 
In Fig. \ref{result1} (A), for the cataract dataset, different colours for phases and steps represent different classes. The first row provides the ground truth (GT) for reference. By examining the timelines of predicted phases and steps, it is evident that our approach exhibits excellent performance even for some challenging classes. For instance, in the step recognition, the S2, highlighted in yellow, constitutes a small proportion of frames in the entire video. Other methods misidentify this category, while our method can accurately predict it. In the action detection task, 
detecting fine-grained actions is more challenging than phase and step recognition, as spatial features for some actions are highly similar, which is also illustrated by the performance of comparison methods. For example, in the left column of Fig. \ref{result1} (A),  cases such as ``lens hook grasp" and ``lens hook hold",
appear visually similar and can be easily misjudged based on appearance alone. Correct judgment necessitates the integration of contextual information, specifically the temporal features of instruments in consecutive frames. 
Notably, our method performs accurately when other approaches struggle to predict the current action, underscoring the effectiveness of our method in learning temporal dynamics. To further demonstrate proposed method's performance in accurately identifying positive instances compared to other methods, we computed the Recall (Re) metrics for all models, as detailed in the supplementary materials. The results demonstrate that our method achieves the best Re performance. 

For the PSI-AVA dataset, as depicted in Fig. \ref{result1} (B), the comparison of phase and step recognition reveals that our method produces prediction results very close to the GT, particularly in regions with dense class distributions, which are also challenging to identify. The action detection task further demonstrates the excellent performance of our method in detecting multiple classes of concurrent actions.


\begin{table}[tp]
\centering
\caption{Ablation study on ICL. The best results are highlighted in bold, while the next best results are in \underline{underline}. }
\label{tabel5}
\setlength{\tabcolsep}{1.5mm}{
    \begin{tabular}{cccccc|cccc}
        \toprule
        \multicolumn{6}{c|}{Items} & \multirow{2}*{Phase}&  \multirow{2}*{Step}&  \multirow{2}*{Instrument}& \multirow{2}*{Action} \\
        \cline{1-6}
         PS& PI & PA& SI& SA& IA&\multicolumn{4}{c}{} \\ \hline
         &  &  &  \checkmark&  \checkmark&  \checkmark&  0.9208&  0.9145&  0.8539&0.5472\\
         &  \checkmark&  \checkmark&  &  &  \checkmark&  \textbf{0.9672}& 0.9113&  \underline{0.8547}&\underline{0.5562}\\ 
         \checkmark&  &  \checkmark&  &  \checkmark&  &  0.9372&  0.8759&  0.8532&0.5481\\
         \checkmark&  \checkmark&  &  \checkmark&  &  &  0.9446&  0.9367&  0.8546&0.5398\\
         \checkmark&  \checkmark&  \checkmark&  \checkmark&  \checkmark&  \checkmark&  0.9353&  0.9240&  \textbf{0.8554}&0.5531\\\hline \hline
         &  &  \checkmark&  \checkmark&  &  &  0.9532&  0.9124& 0.8543&0.5535\\
         &  \checkmark&  &  &  \checkmark&  &  0.9580&  \underline{0.9452}&  0.8544&0.5478\\
         \checkmark&  &  &  &  &  \checkmark& \underline{0.9635}&\textbf{0.9581}&   0.8545&\textbf{0.5726}\\
         \bottomrule
    \end{tabular}}
\end{table}


\subsection{Ablation Study}
\subsubsection{Contribution of Key Model Components}
We conduct ablation experiments on two datasets to validate the contribution of key modules, including HRAM and ICL, with the results presented in Table \ref{tabel3}. The models excluding the HRAM and ICL modules serve as our baseline models. It is built upon the MViTv2 framework, supplemented with multi-task loss and multi-task output heads. 
To minimize computational costs, all of our improvements are based on the final block. We compare the effects of adding HRAM to the baseline model, as well as the combination of HRAM and ICL modules.

For the cataract dataset, adding HRAM improves the performance across all tasks. It enhances phase recognition, step recognition and action detection by 0.29\% (95.23\% vs. 94.94\%), 2.00\% (94.89\% vs. 92.89\%) and 1.19\% (56.82\% vs. 55.63\%), respectively.  This indicates a strong correlation and dependency among the four tasks within the cataract dataset. After further incorporating the ICL module, the performance of various tasks is further enhanced. This demonstrates that task-specific features are strengthened through intrinsic consistency extracted by the ICL module. The performance drop in instrument detection after incorporating the ICL module stems from limited task complementarity, as instrument detection, which depends on spatial precision, benefits less from shared temporal information across other tasks. Additionally, the multi-task learning framework is designed to prioritize overall task balance. Nonetheless, the results demonstrate that these effects are minimal in our model. 

In parallel, for the PSI-AVA dataset, due to substantial differences in the distribution of labels for different tasks as described above, our experimental design follows \cite{valderrama2022towards} to implement the four tasks. This means that we have to divide the experiments into two parts: phase and step, and instrument and action. Consequently, the HRAM and ICL modules only achieve mutual enhancement between two tasks, rather than incorporating information from all four tasks.
However, even under these constraints, the addition of the HRAM still results in superior performance compared to the baseline model. Further improvements are observed with the addition of the ICL module, with an improvement of even 2.68\% (64.57\% vs. 61.89\%) in phase recognition against the baseline model.

t-SNE visualization results illustrating the contribution of our key model components are depicted in Fig. \ref{tsne}. The blue circle in Fig. \ref{tsne} demonstrates that adding our key modules improves the distinguishability of classes and reduces inter-class blurring across phase, step, and action tasks. For the action detection task, due to space constraints, we focus on 7 classes generated by the instrumented cannula. These classes exhibit significant spatial similarity, posing classification challenges. The baseline model mixes these classes as shown in Fig. \ref{tsne} (c). Upon integrating our proposed module, the distinctiveness among these classes, highlighted by the blue circle, notably increases. To quantitatively validate this observation, we compare the Average Precision (AP) values of these seven classes across three models in Fig. \ref{bar}, aligning with their labeling in Fig. \ref{tsne}, thus confirming the efficacy of our proposed module.

\subsubsection{Contribution of HRAM}
For a more detailed analysis of the HRAM module's role in enhancing interactions among various tasks, we conduct experimental analyses as presented in Table  \ref{tabel4}. ``P", ``S", ``I", and ``A" represent the tasks of phase, step, instrument, and action respectively. A checkmark beside a task indicates its participation in mutual enhancement, while its absence signifies exclusion. 
This approach allows us to analyze the impact of each task on the performance of a multi-task framework.

Table \ref{tabel4} reveals that when the influence of ``P" is excluded, the mAP for phase recognition is relatively lower (91.53\%). When ``S" is excluded, the performance of step recognition is the poorest (only 87.39\%). A similar pattern is observed in the action detection task. This suggests that the optimization of a task is significantly diminished when the task lacks participation in interactions with other tasks. This also underscores the importance of other tasks in optimizing or enhancing the specific task. Notably, when instrument features are missing, the performance of all phase, step, and action tasks is comparatively poor, highlighting the critical guiding role of instrument features for these three tasks. This aligns with our understanding of surgical videos, where certain specific instruments can directly indicate the current phase and step of a frame, and even some simple ongoing actions can be inferred. The last row of the table shows that when all tasks are incorporated into our framework and mutually enhance each other, we achieve optimal performance for all four tasks with even mAP of 95.23\% and 94.89\% for phase and step.
\begin{figure}[bp]
\begin{center}
\includegraphics[width=\linewidth]{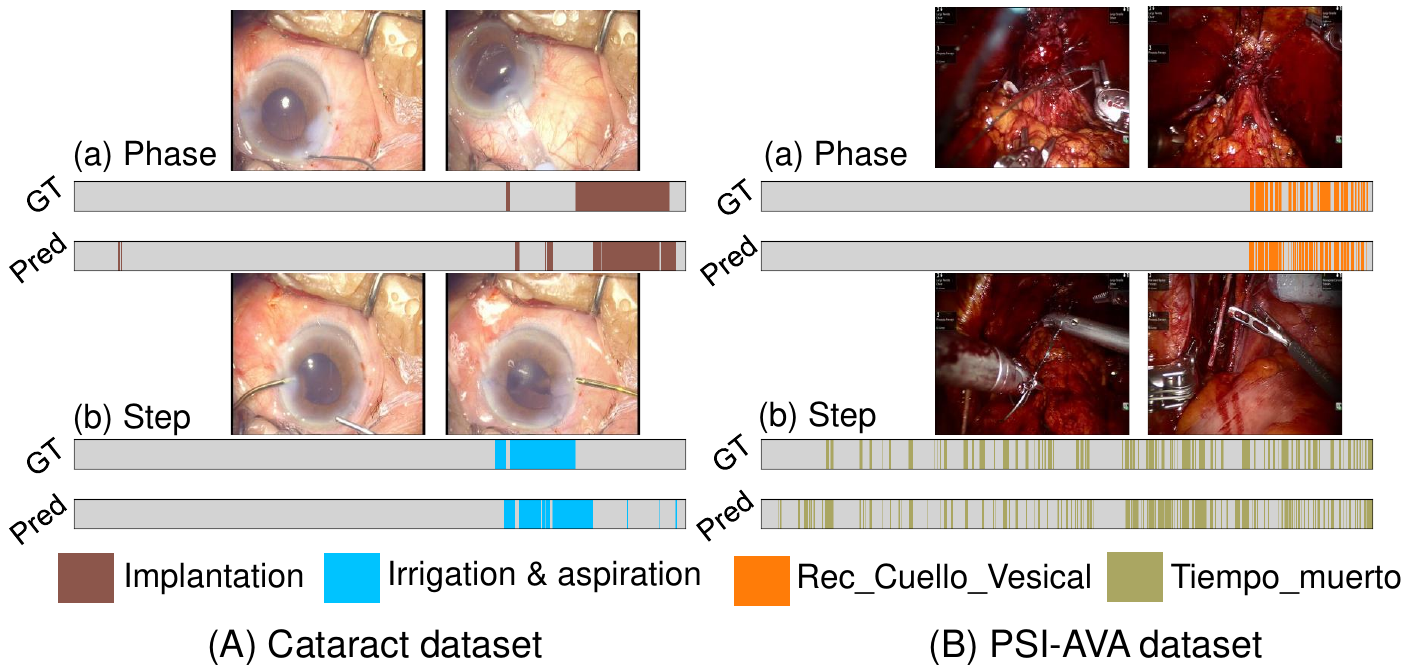}

\caption{Color-coded ribbon visualization of phase and step recognition for some challenging classes from the proposed method (Pred) and the ground truth (GT) in two complete surgical videos. In each case, we show the classes with different colors and no class appearance with blank. Horizontal axes indicate the time progression.}

\label{difficult}
\end{center}
\end{figure}

\subsubsection{Contribution of ICL}
To verify the effectiveness of ICL, we carry on detailed ablation experiments based on the cataract dataset. Since ICL optimizes the model by calculating losses for pairwise task features, we design ablation experiments in two aspects (separated by double solid lines in Table \ref{tabel5}) to assess the performance differences of each task within ICL.  The table lists six pairs of our four tasks, namely ``PS'', ``PI'', ``PA'', ``SI'', ``SA'', and ``IA''. On one hand, similar to HRAM, we calculate ICL by excluding the influence of one task.  Table  \ref{tabel5} (above the double solid lines) shows the model's performance when calculating ICL pairwise among the remaining three tasks after excluding one task (``P'', ``S'', ``I'', or ``A''). 
We observe that relatively lower results are yielded for a task when it is excluded by ICL. In Table  \ref{tabel5}, excluding ``P'' leads to the poorest performance in the phase recognition (only 92.08\%). Excluding ``S'' results in the second-worst performance in step recognition. The absence of ``I'' leads to lower performance in phase and step recognition, as well as instrument detection. The mAP for step recognition even drops to only 87.59\%. Similarly, excluding ``A'' reduces the performance of action detection. This confirms the importance of each task in the functioning of ICL.

However, due to inherent differences between tasks, the optimal results are not able to be yielded when all four task pairs are fed into ICL, as shown in Table \ref{tabel5}. In our tasks, phase and step have similar optimization objectives and can extract the strongest consistencies for the same positive sample, as do action and instrument. Our second set of experiments, as illustrated in Table  \ref{tabel5} (below the double solid lines), confirms this. When we optimize the model by combining the pairs of ``PS'' and ``IA'', the model achieves optimal performance in each task, further improving upon the addition of HRAM.

\subsection{Effectiveness of HCT+}

Utilizing the Kinetics\-400 \cite{carreira2017quo} pretrained model provided by MViTv2 \cite{li2022mvitv2}, we conduct experiments to validate the effectiveness of the ST-Ada, as presented in Table \ref{tabel6}. 
The peak memory is measured when the batch size is 16. ``Full-tuning" indicates fine-tuning of the entire model.
As a reference baseline method, we perform full fine-tuning based on the pretrained model of the MViTv2 framework, which gets the poorest performance. When fully fine-tuning our proposed model, the performance on the four tasks reaches the optimum, but the ``param'' is also the highest at 57.2M, and memory consumption is at its peak. If we freeze the parameters of the first 15 transformer blocks of our proposed model and only optimize the last improved block, denoted as ``w/o ST\_Ada'', the number of parameters to be optimized decreases by 57.6\%. However, the performance on each task also decreases, with a notable drop of 4.60\% (91.75\% vs. 96.35\%) in phase recognition. Nevertheless, with the addition of the spatial adapter (w/S\_Ada), the model's performance experiences a widespread improvement, with only a marginal increase of 1.2\% in parameters based on the ``w/o ST\_Ada'' model. Additionally, during training, peak memory decreases by nearly 10G compared to our full-tuning model, significantly reducing computational costs. Similarly, when only the temporal adapter is added (w/T\_Ada), with a minimal increase of 9.5\% in parameters, the performance on each task is better than without any adapters. This indicates the excellent transferability of spatial and temporal adapters. Upon combining spatial and temporal adapters (HCT+), our model's performance further improves with only a 10.7\% increase in parameters. The accuracy of phase and step recognition even surpasses that of ``w/o ST\_Ada'' by 3.67\% (95.42\% vs. 91.75\%) and 4.70\% (94.37\% vs. 89.67\%), respectively. Here is just a preliminary exploration of the application of transfer learning in our task. Its feasibility further makes it possible to apply our multi-level semantic scene understanding task facts to surgical scenarios.

\begin{table}[tp]
    \centering
\caption{Performance on the cataract dataset for HCT+. ``Param" denotes the number of parameters and ``PM" denotes peak memory usage. }
\label{tabel6}
\setlength{\tabcolsep}{0.2mm}{
    \begin{tabular}{c|cc|cccc} 
    \toprule
        Method&Params(M) &PM(MB)& Phase&  Step&   Instrument& Action   \\\hline
          MViTv2(Full-tuning)& 35.7 &34382 & 0.9494& 0.9289 & 0.8543 & 0.5563  \\
          HCT(Full-tuning) &  57.2 (100\%) &42850 &  \textbf{0.9635}& \textbf{ 0.9581}& 0.8545&  \textbf{0.5726}\\
          HCT(w/o ST\_Ada)&  24.3 (42.4\%)&32816 &  0.9175&   0.8967& 0.8541&0.5412\\\hline\hline
          HCT(w/ S\_Ada)&  24.6 (+1.2\%) & 32824 &   0.9381&  0.9301& 0.8539&  0.5613\\
          HCT(w/ T\_Ada)&  26.6 (+9.5\%)&35610 &  0.9345& 0.9271  &  \underline{0.8545}&  0.5537\\  
 HCT+(w/ ST\_Ada)& 26.9 (+10.7\%) & 35616 & \underline{0.9542} &\underline{0.9437} &\textbf{0.8546} & \underline{0.5672}  \\ \bottomrule   
    \end{tabular}}    
\end{table}

\subsection{Challenge Cases Analysis}
We demonstrate the recognition performance of our model across phases and steps for challenging classes, as shown in Fig. \ref{difficult}. The classes are represented in different colors, with blank spaces indicating no class appearance. For the cataract dataset, as depicted in the video frame in Fig. \ref{difficult} (A).a, it is difficult to extract effective spatial information for discrimination. This is because the “Cannula” in the left figure is at the edge of the field of view, and the ``Implant injector" in the right figure is transparent. In Fig. \ref{difficult} (A).b, the incomplete display of the ``Handpiece" head instrument can lead to confusion with similar instruments, such as “Cannula”, posing a classification challenge.

In the PSI-AVA dataset, both the phase and step class distribution shown in Fig. \ref{difficult} B are scattered, making it challenging to differentiate classes using temporal information. In Fig. \ref{difficult} (B).a, the confusion between instruments and tissue in the video frame complicates the distinction of operations. In Fig. \ref{difficult} (B).b, the step ``Tiempo\_muerto" indicates the absence of a step, which is widely dispersed and scattered in the whole video, making it more difficult to identify null operations compared to other operations. Despite these challenges, our method predicts a distribution very similar to the ground truth (GT). Combined with the results shown in Fig. \ref{bar}, this demonstrates the robustness and effectiveness of our method for all tasks.

\section{Conclusion}
In this paper, we present a novel hierarchical context transformer (HCT) network to access MSSU by achieving recognition of phases and steps, and detection of actions and instruments. Specifically, we develop a hierarchical relation aggregation module (HRAM) to dig out the intrinsic relationships between different tasks and further boost task-wise representation learning with the proposed inter-task contrastive learning (ICL). For the costs of model training and storage on the transformer models, we introduce spatial and temporal adapters to equip our HCT for HCT+ with spatial-temporal reasoning capability, which enables the HCT+ to achieve comparable performance with the full fine-tuned model at a small parameter cost.
Comprehensive experimental results on two surgical video datasets demonstrate the superiority of our proposed model and the contribution of each key component.

\section{Acknowledgement}
This work was supported in part by General Program of National Natural Science Foundation of China (82102189 and 82272086).

\bibliographystyle{ieeetr}
\bibliography{HCT}

\begin{thebibliography}{10}

\bibitem{vercauteren2019cai4cai}
T.~Vercauteren, M.~Unberath, N.~Padoy, and N.~Navab, ``Cai4cai: the rise of contextual artificial intelligence in computer-assisted interventions,'' {\em Proceedings of the IEEE}, vol.~108, no.~1, pp.~198--214, 2019.

\bibitem{lalys2014surgical}
F.~Lalys and P.~Jannin, ``Surgical process modelling: a review,'' {\em International journal of computer assisted radiology and surgery}, vol.~9, pp.~495--511, 2014.

\bibitem{hao2023act}
L.~Hao, Y.~Hu, W.~Lin, Q.~Wang, H.~Li, H.~Fu, J.~Duan, and J.~Liu, ``Act-net: Anchor-context action detection in surgery videos,'' in {\em International Conference on Medical Image Computing and Computer-Assisted Intervention}, pp.~196--206, Springer, 2023.

\bibitem{huaulme2021micro}
A.~Huaulm{\'e}, D.~Sarikaya, K.~Le~Mut, F.~Despinoy, Y.~Long, Q.~Dou, C.-B. Chng, W.~Lin, S.~Kondo, L.~Bravo-S{\'a}nchez, {\em et~al.}, ``Micro-surgical anastomose workflow recognition challenge report,'' {\em Computer Methods and Programs in Biomedicine}, vol.~212, p.~106452, 2021.

\bibitem{seenivasan2023surgicalgpt}
L.~Seenivasan, M.~Islam, G.~Kannan, and H.~Ren, ``Surgicalgpt: end-to-end language-vision gpt for visual question answering in surgery,'' in {\em International conference on medical image computing and computer-assisted intervention}, pp.~281--290, Springer, 2023.

\bibitem{wang2023truncate}
Z.~Wang, J.~Weng, C.~Yuan, and J.~Wang, ``Truncate-split-contrast: a framework for learning from mislabeled videos,'' in {\em Proceedings of the AAAI Conference on Artificial Intelligence}, vol.~37, pp.~2751--2758, 2023.

\bibitem{huaulme2020offline}
A.~Huaulm{\'e}, P.~Jannin, F.~Reche, J.-L. Faucheron, A.~Moreau-Gaudry, and S.~Voros, ``Offline identification of surgical deviations in laparoscopic rectopexy,'' {\em Artificial Intelligence in Medicine}, vol.~104, p.~101837, 2020.

\bibitem{zhang2023surgical}
B.~Zhang, B.~Goel, M.~H. Sarhan, V.~K. Goel, R.~Abukhalil, B.~Kalesan, N.~Stottler, and S.~Petculescu, ``Surgical workflow recognition with temporal convolution and transformer for action segmentation,'' {\em International Journal of Computer Assisted Radiology and Surgery}, vol.~18, no.~4, pp.~785--794, 2023.

\bibitem{mascagni2022artificial}
P.~Mascagni, A.~Vardazaryan, D.~Alapatt, T.~Urade, T.~Emre, C.~Fiorillo, P.~Pessaux, D.~Mutter, J.~Marescaux, G.~Costamagna, {\em et~al.}, ``Artificial intelligence for surgical safety: automatic assessment of the critical view of safety in laparoscopic cholecystectomy using deep learning,'' {\em Annals of surgery}, vol.~275, no.~5, pp.~955--961, 2022.

\bibitem{sun2023masked}
X.~Sun, P.~Chen, L.~Chen, C.~Li, T.~H. Li, M.~Tan, and C.~Gan, ``Masked motion encoding for self-supervised video representation learning,'' in {\em Proceedings of the IEEE/CVF Conference on Computer Vision and Pattern Recognition}, pp.~2235--2245, 2023.

\bibitem{padoy2019machine}
N.~Padoy, ``Machine and deep learning for workflow recognition during surgery,'' {\em Minimally Invasive Therapy \& Allied Technologies}, vol.~28, no.~2, pp.~82--90, 2019.

\bibitem{liu2021towards}
D.~Liu, Q.~Li, T.~Jiang, Y.~Wang, R.~Miao, F.~Shan, and Z.~Li, ``Towards unified surgical skill assessment,'' in {\em Proceedings of the IEEE/CVF Conference on Computer Vision and Pattern Recognition}, pp.~9522--9531, 2021.

\bibitem{zhou2023hierarchical}
K.~Zhou, Y.~Ma, H.~P. Shum, and X.~Liang, ``Hierarchical graph convolutional networks for action quality assessment,'' {\em IEEE Transactions on Circuits and Systems for Video Technology}, vol.~33, no.~12, pp.~7749--7763, 2023.

\bibitem{li2024continual}
Y.-M. Li, L.-A. Zeng, J.-K. Meng, and W.-S. Zheng, ``Continual action assessment via task-consistent score-discriminative feature distribution modeling,'' {\em IEEE Transactions on Circuits and Systems for Video Technology}, 2024.

\bibitem{yue2023perceptual}
G.~Yue, D.~Cheng, T.~Zhou, J.~Hou, W.~Liu, L.~Xu, T.~Wang, and J.~Cheng, ``Perceptual quality assessment of enhanced colonoscopy images: A benchmark dataset and an objective method,'' {\em IEEE Transactions on Circuits and Systems for Video Technology}, vol.~33, no.~10, pp.~5549--5561, 2023.

\bibitem{jin2021temporal}
Y.~Jin, Y.~Long, C.~Chen, Z.~Zhao, Q.~Dou, and P.-A. Heng, ``Temporal memory relation network for workflow recognition from surgical video,'' {\em IEEE Transactions on Medical Imaging}, vol.~40, no.~7, pp.~1911--1923, 2021.

\bibitem{czempiel2020tecno}
T.~Czempiel, M.~Paschali, M.~Keicher, W.~Simson, H.~Feussner, S.~T. Kim, and N.~Navab, ``Tecno: Surgical phase recognition with multi-stage temporal convolutional networks,'' in {\em International Conference on Medical Image Computing and Computer-Assisted Intervention}, pp.~343--352, Springer, 2020.

\bibitem{yi2022not}
F.~Yi, Y.~Yang, and T.~Jiang, ``Not end-to-end: Explore multi-stage architecture for online surgical phase recognition,'' in {\em Proceedings of the Asian Conference on Computer Vision}, pp.~2613--2628, 2022.

\bibitem{pan2023temporal}
X.~Pan, X.~Gao, H.~Wang, W.~Zhang, Y.~Mu, and X.~He, ``Temporal-based swin transformer network for workflow recognition of surgical video,'' {\em International Journal of Computer Assisted Radiology and Surgery}, vol.~18, no.~1, pp.~139--147, 2023.

\bibitem{twinanda2016endonet}
A.~P. Twinanda, S.~Shehata, D.~Mutter, J.~Marescaux, M.~De~Mathelin, and N.~Padoy, ``Endonet: a deep architecture for recognition tasks on laparoscopic videos,'' {\em IEEE transactions on medical imaging}, vol.~36, no.~1, pp.~86--97, 2016.

\bibitem{czempiel2021opera}
T.~Czempiel, M.~Paschali, D.~Ostler, S.~T. Kim, B.~Busam, and N.~Navab, ``Opera: Attention-regularized transformers for surgical phase recognition,'' in {\em International Conference on Medical Image Computing and Computer-Assisted Intervention}, pp.~604--614, Springer, 2021.

\bibitem{ding2022exploring}
X.~Ding and X.~Li, ``Exploring segment-level semantics for online phase recognition from surgical videos,'' {\em IEEE Transactions on Medical Imaging}, vol.~41, no.~11, pp.~3309--3319, 2022.

\bibitem{yue2023cascade}
W.~Yue, H.~Liao, Y.~Xia, V.~Lam, J.~Luo, and Z.~Wang, ``Cascade multi-level transformer network for surgical workflow analysis,'' {\em IEEE Transactions on Medical Imaging}, 2023.

\bibitem{nwoye2022rendezvous}
C.~I. Nwoye, T.~Yu, C.~Gonzalez, B.~Seeliger, P.~Mascagni, D.~Mutter, J.~Marescaux, and N.~Padoy, ``Rendezvous: Attention mechanisms for the recognition of surgical action triplets in endoscopic videos,'' {\em Medical Image Analysis}, vol.~78, p.~102433, 2022.

\bibitem{lin2022instrument}
W.~Lin, Y.~Hu, L.~Hao, D.~Zhou, M.~Yang, H.~Fu, C.~Chui, and J.~Liu, ``Instrument-tissue interaction quintuple detection in surgery videos,'' in {\em International Conference on Medical Image Computing and Computer-Assisted Intervention}, pp.~399--409, Springer, 2022.

\bibitem{xi2022forest}
N.~Xi, J.~Meng, and J.~Yuan, ``Forest graph convolutional network for surgical action triplet recognition in endoscopic videos,'' {\em IEEE Transactions on Circuits and Systems for Video Technology}, vol.~32, no.~12, pp.~8550--8561, 2022.

\bibitem{wang2023vision}
W.~Wang, X.~Yang, and J.~Tang, ``Vision transformer with hybrid shifted windows for gastrointestinal endoscopy image classification,'' {\em IEEE Transactions on Circuits and Systems for Video Technology}, vol.~33, no.~9, pp.~4452--4461, 2023.

\bibitem{liu2024xfmp}
J.~Liu, J.~Niu, W.~Li, X.~Li, B.~He, H.~Zhou, Y.~Liu, D.~Li, B.~Wang, and W.~Zhang, ``Xfmp: A benchmark for explainable fine-grained abnormal behavior recognition on medical personal protective equipment,'' {\em IEEE Transactions on Circuits and Systems for Video Technology}, 2024.

\bibitem{valderrama2022towards}
N.~Valderrama, P.~Ruiz~Puentes, I.~Hern{\'a}ndez, N.~Ayobi, M.~Verlyck, J.~Santander, J.~Caicedo, N.~Fern{\'a}ndez, and P.~Arbel{\'a}ez, ``Towards holistic surgical scene understanding,'' in {\em International Conference on Medical Image Computing and Computer-Assisted Intervention}, pp.~442--452, Springer, 2022.

\bibitem{zhang2022unsupervised}
C.~Zhang, T.~Yang, J.~Weng, M.~Cao, J.~Wang, and Y.~Zou, ``Unsupervised pre-training for temporal action localization tasks,'' in {\em Proceedings of the IEEE/CVF conference on computer vision and pattern recognition}, pp.~14031--14041, 2022.

\bibitem{lou2023min}
A.~Lou, K.~Tawfik, X.~Yao, Z.~Liu, and J.~Noble, ``Min-max similarity: A contrastive semi-supervised deep learning network for surgical tools segmentation,'' {\em IEEE Transactions on Medical Imaging}, 2023.

\bibitem{pan2022st}
J.~Pan, Z.~Lin, X.~Zhu, J.~Shao, and H.~Li, ``St-adapter: Parameter-efficient image-to-video transfer learning,'' {\em Advances in Neural Information Processing Systems}, vol.~35, pp.~26462--26477, 2022.

\bibitem{yang2023aim}
T.~Yang, Y.~Zhu, Y.~Xie, A.~Zhang, C.~Chen, and M.~Li, ``{AIM}: Adapting image models for efficient video action recognition,'' in {\em The Eleventh International Conference on Learning Representations}, 2023.

\bibitem{bawa2021saras}
V.~S. Bawa, G.~Singh, F.~KapingA, I.~Skarga-Bandurova, E.~Oleari, A.~Leporini, C.~Landolfo, P.~Zhao, X.~Xiang, G.~Luo, {\em et~al.}, ``The saras endoscopic surgeon action detection (esad) dataset: Challenges and methods,'' {\em arXiv preprint arXiv:2104.03178}, 2021.

\bibitem{ren2015faster}
S.~Ren, K.~He, R.~Girshick, and J.~Sun, ``Faster r-cnn: Towards real-time object detection with region proposal networks,'' {\em Advances in neural information processing systems}, vol.~28, 2015.

\bibitem{ramesh2021multi}
S.~Ramesh, D.~Dall’Alba, C.~Gonzalez, T.~Yu, P.~Mascagni, D.~Mutter, J.~Marescaux, P.~Fiorini, and N.~Padoy, ``Multi-task temporal convolutional networks for joint recognition of surgical phases and steps in gastric bypass procedures,'' {\em International journal of computer assisted radiology and surgery}, vol.~16, pp.~1111--1119, 2021.

\bibitem{nwoye2020recognition}
C.~I. Nwoye, C.~Gonzalez, T.~Yu, P.~Mascagni, D.~Mutter, J.~Marescaux, and N.~Padoy, ``Recognition of instrument-tissue interactions in endoscopic videos via action triplets,'' in {\em International Conference on Medical Image Computing and Computer-Assisted Intervention}, pp.~364--374, Springer, 2020.

\bibitem{lin2024instrument}
W.~Lin, Y.~Hu, H.~Fu, M.~Yang, C.-B. Chng, R.~Kawasaki, C.~Chui, and J.~Liu, ``Instrument-tissue interaction detection framework for surgical video understanding,'' {\em IEEE Transactions on Medical Imaging}, 2024.

\bibitem{li2022mvitv2}
Y.~Li, C.-Y. Wu, H.~Fan, K.~Mangalam, B.~Xiong, J.~Malik, and C.~Feichtenhofer, ``Mvitv2: Improved multiscale vision transformers for classification and detection,'' in {\em Proceedings of the IEEE/CVF Conference on Computer Vision and Pattern Recognition}, pp.~4804--4814, 2022.

\bibitem{fan2021multiscale}
H.~Fan, B.~Xiong, K.~Mangalam, Y.~Li, Z.~Yan, J.~Malik, and C.~Feichtenhofer, ``Multiscale vision transformers,'' in {\em Proceedings of the IEEE/CVF international conference on computer vision}, pp.~6824--6835, 2021.

\bibitem{zhang2023dino}
H.~Zhang, F.~Li, S.~Liu, L.~Zhang, H.~Su, J.~Zhu, L.~Ni, and H.-Y. Shum, ``{DINO}: {DETR} with improved denoising anchor boxes for end-to-end object detection,'' in {\em The Eleventh International Conference on Learning Representations}, 2023.

\bibitem{houlsby2019parameter}
N.~Houlsby, A.~Giurgiu, S.~Jastrzebski, B.~Morrone, Q.~De~Laroussilhe, A.~Gesmundo, M.~Attariyan, and S.~Gelly, ``Parameter-efficient transfer learning for nlp,'' in {\em International Conference on Machine Learning}, pp.~2790--2799, PMLR, 2019.

\bibitem{feichtenhofer2019slowfast}
C.~Feichtenhofer, H.~Fan, J.~Malik, and K.~He, ``Slowfast networks for video recognition,'' in {\em Proceedings of the IEEE/CVF international conference on computer vision}, pp.~6202--6211, 2019.

\bibitem{carreira2017quo}
J.~Carreira and A.~Zisserman, ``Quo vadis, action recognition? a new model and the kinetics dataset,'' in {\em proceedings of the IEEE Conference on Computer Vision and Pattern Recognition}, pp.~6299--6308, 2017.

\bibitem{allan20192017}
M.~Allan, A.~Shvets, T.~Kurmann, Z.~Zhang, R.~Duggal, Y.-H. Su, N.~Rieke, I.~Laina, N.~Kalavakonda, S.~Bodenstedt, {\em et~al.}, ``2017 robotic instrument segmentation challenge,'' {\em arXiv preprint arXiv:1902.06426}, 2019.

\bibitem{allan20202018}
M.~Allan, S.~Kondo, S.~Bodenstedt, S.~Leger, R.~Kadkhodamohammadi, I.~Luengo, F.~Fuentes, E.~Flouty, A.~Mohammed, M.~Pedersen, {\em et~al.}, ``2018 robotic scene segmentation challenge,'' {\em arXiv preprint arXiv:2001.11190}, 2020.

\end{thebibliography}

\begin{IEEEbiography}
[{\includegraphics[width=1in,height=1.25in,clip,keepaspectratio]{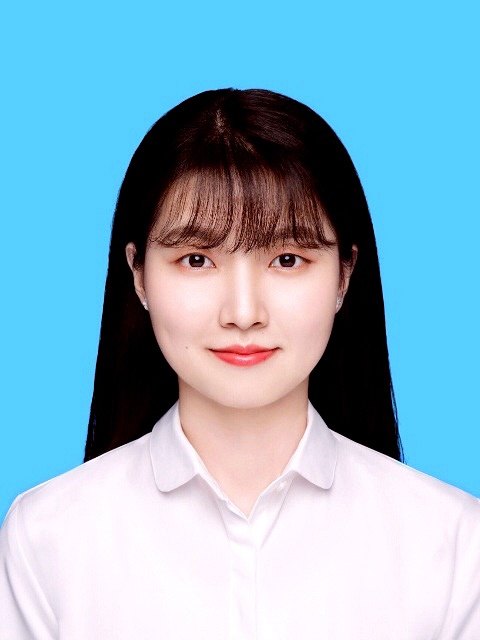}}]{Luoying Hao}
 is a PhD student of the School of Computer Science from the University of Birmingham, UK. His research interests include medical image analysis, surgical video analysis and scene understanding.
\end{IEEEbiography}
\vspace{-20 pt}
\begin{IEEEbiography}[{\includegraphics[width=1in,height=1.25in,clip,keepaspectratio]{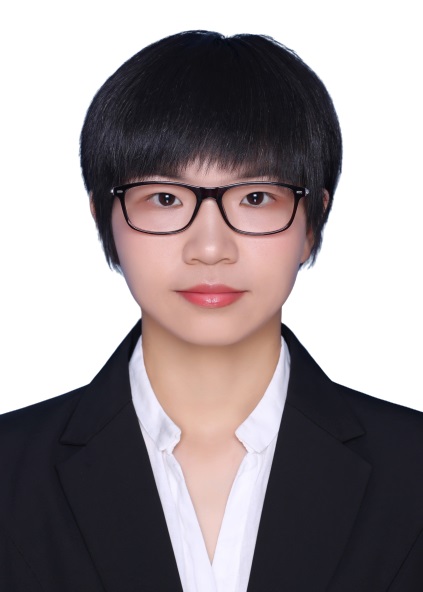}}]{Yan Hu}
 received the PhD degree from the Department of Information Science and Technology, the University of Tokyo, Japan. She is working now in the Southern University of Science and Technology, China. Her research interests include medical image analysis, surgery video processing, and computer-aided surgery.
\end{IEEEbiography}
 \vspace{-20 pt}
\begin{IEEEbiography}
[{\includegraphics[width=1in,height=1.25in,clip,keepaspectratio]{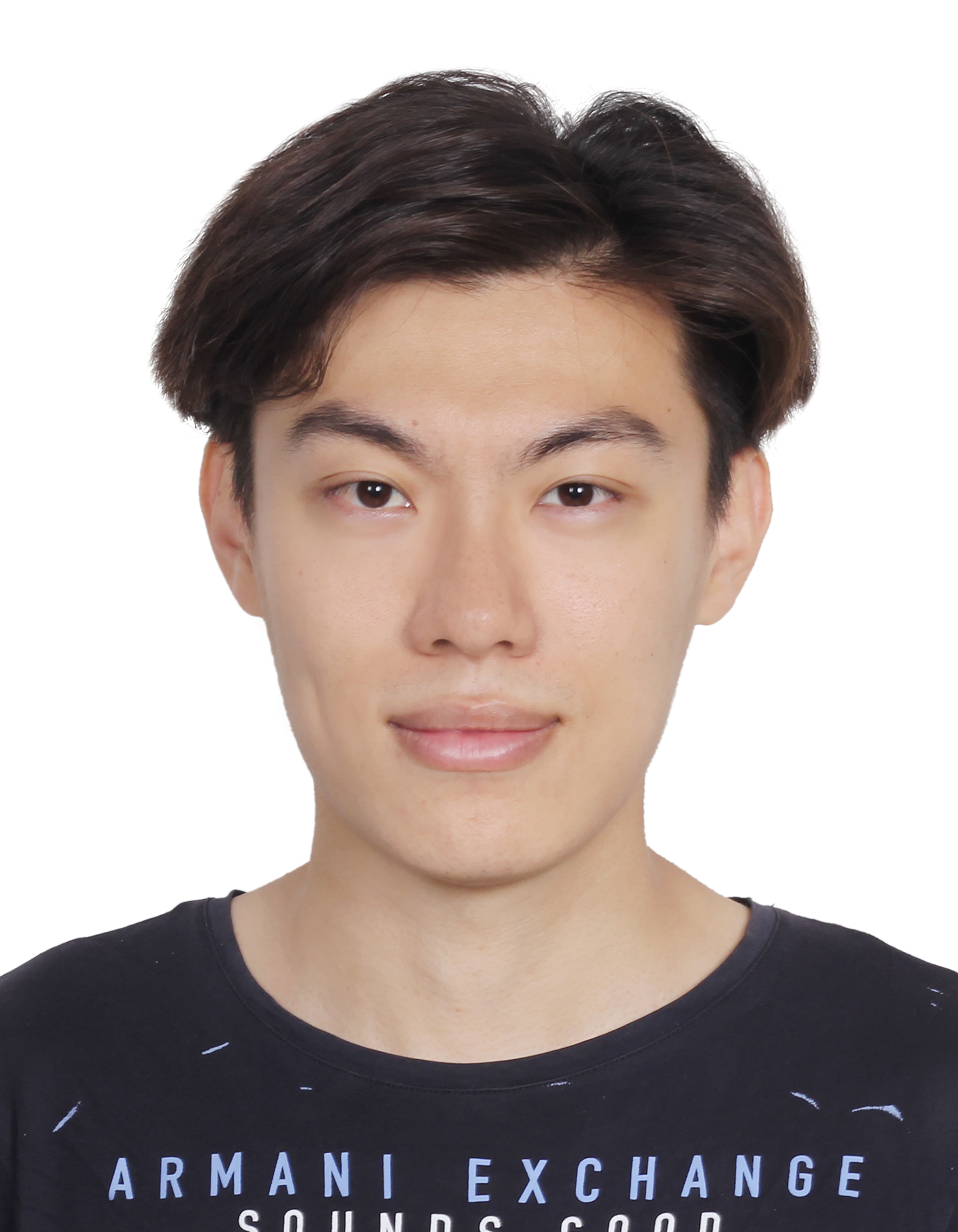}}]{Yang Yue}
 is a PhD student of the School of Computer Science from the University of Birmingham, UK. His research interests include bioinformatics, machine learning and data mining.
\end{IEEEbiography}
\vspace{-20 pt}
\begin{IEEEbiography}[{\includegraphics[width=1in,height=1.25in,clip,keepaspectratio]{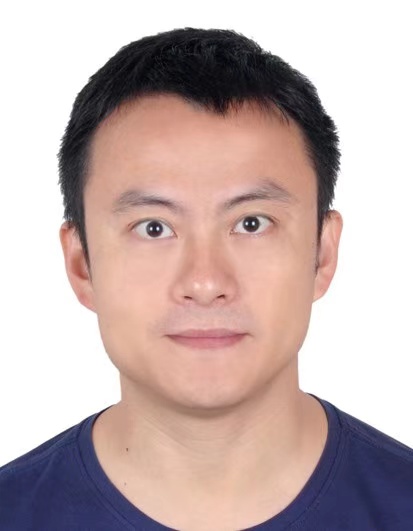}}]{Li Wu}
 received  the Master’s degree in Measuring and Testing Technologies and Instruments from the University of Electronic Science and Technology of China in 2004. He is currently the general manager of Cloud Shadow Medical Technology Co., Ltd. at BGI Genomics. With over 20 years of experience in the medical device industry, he has held positions at Mindray, BGI, and BGI Genomics. He has been involved in and led the development of high-end medical devices with independent intellectual property rights, including digital ultrasound, high-precision infusion pumps, high-throughput sequencers, automated library preparation equipment, remote ultrasound robots, and automated ultrasound robots. His research interests include upper and lower computer communication, information systems, robotic motion control, and medical imaging AI.
\end{IEEEbiography}
\vspace{-20 pt}
\begin{IEEEbiography}
[{\includegraphics[width=1in,height=1.25in,clip,keepaspectratio]{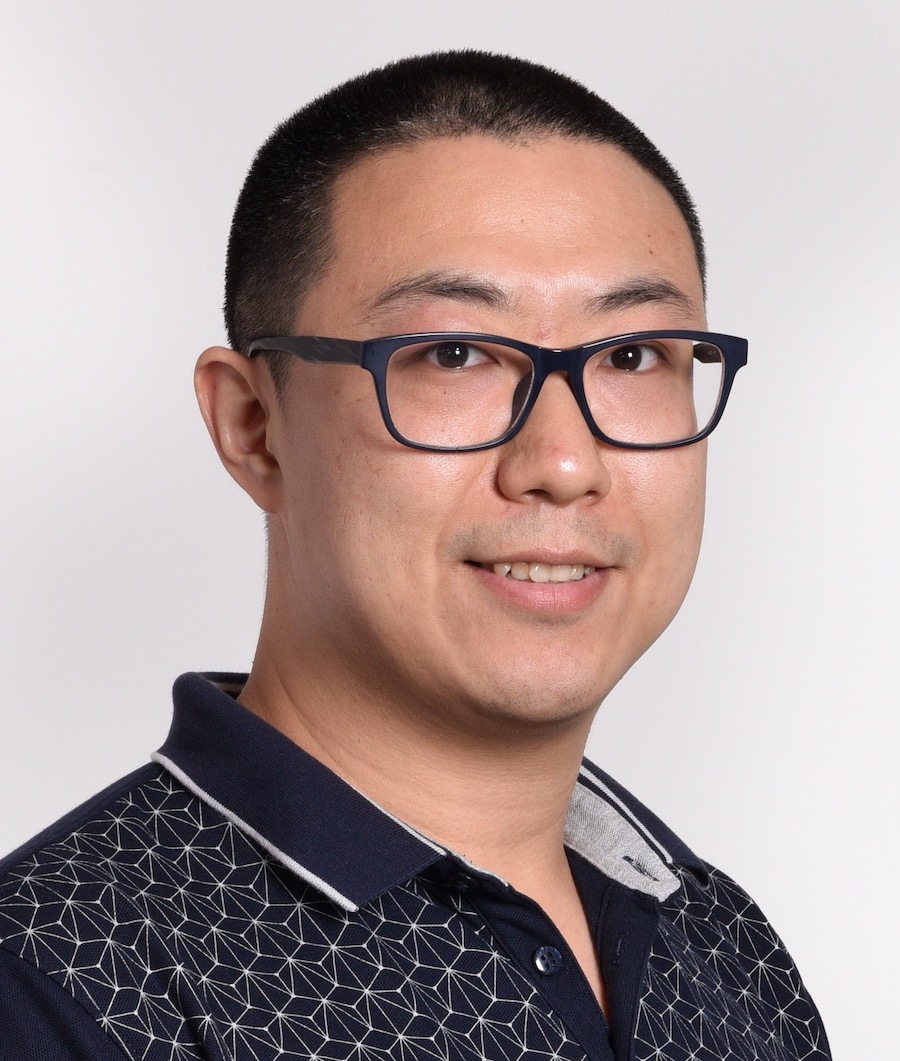}}]{Huazhu Fu}
(Senior Member, IEEE) received the
Ph.D. degree from Tianjin University, Tianjin, China,
in 2013. He is currently a senior Scientist with IHPC,
A*STAR, Singapore. He was a Research Fellow with
Nanyang Technological University (NTU), Singapore, during 2013–2015, a Research Scientist with
the Institute for Infocomm Research (I2R), A*STAR,
Singapore, during 2015–2018, and a Senior Scientist with Inception Institute of Artificial Intelligence
(IIAI), UAE, during 2018–2021. His research interests include computer vision, AI in healthcare, and
trustworthy AI.
\end{IEEEbiography}
\vspace{-20 pt}
\begin{IEEEbiography}
[{\includegraphics[width=1in,height=1.25in,clip,keepaspectratio]{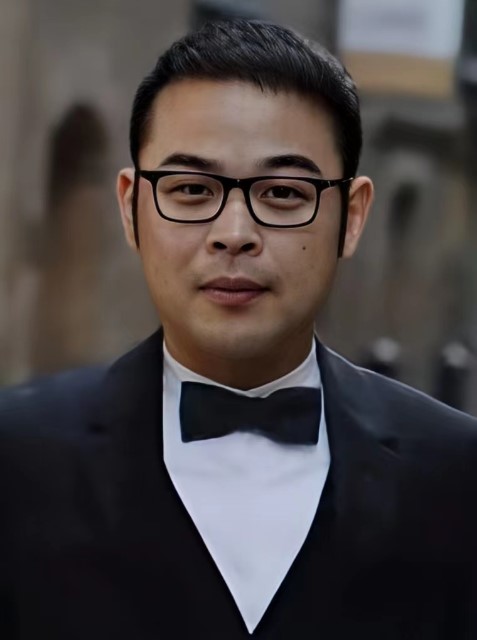}}]{Jinming Duan}
received the Ph.D. degree in computer science from the University of Nottingham.
He is currently a Turing Fellow at the Alan Turing
Institute and an Assistant Professor of computer
science at the University of Birmingham. He is also
a fellow of the Higher Education Academy (FHEA)
under the U.K. Professional Standards Framework
for teaching and learning support in higher education. He was a Research Associate jointly at the
Department of Computing, Institute of Clinical Sciences, Imperial College London, where he has been
developed cutting-edge machine learning methods for cardiovascular imaging
problems. His Ph.D. was funded by the Engineering and Physical Sciences
Research Council. His research interests include deep neural nets, variational
methods, partial/ordinary differential equations, numerical optimisation, and
finite difference/element methods, with applications to image processing,
computer vision, and medical imaging analysis.
\end{IEEEbiography}
\vspace{-20 pt}
\begin{IEEEbiography}
[{\includegraphics[width=1in,height=1.25in,clip,keepaspectratio]{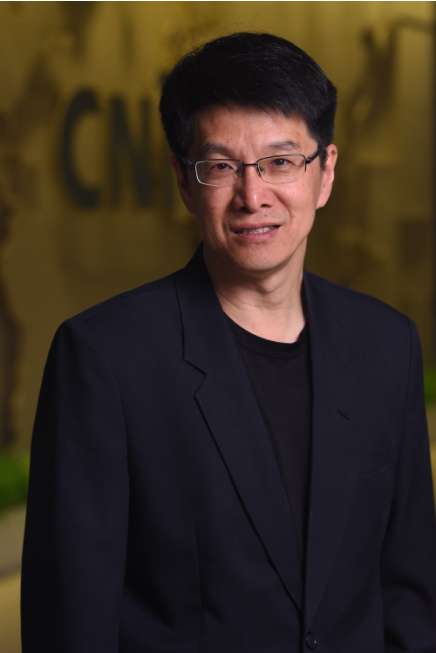}}]{Jiang Liu}
(Senior Member, IEEE) received the bachelor’s degree from the Department of Computer Science, University of Science and Technology of China,
Hefei, China, and the master’s and doctorate degree
from the Department of Computer Science, National
University of Singapore, Singapore. He is currently
a tenured Professor with Southern University of Science and Technology and Founder of the iMED China
team. His research interests include artificial intelligence in ophthalmology, eye-brain linkage, precision
medicine, and surgical robots.
\end{IEEEbiography}



\end{document}